\documentclass[12pt]{article}
\usepackage{amsmath, amsthm, amssymb}
\usepackage{graphicx}
\usepackage{enumerate}
\usepackage{multirow}
\usepackage[sort]{natbib}
\usepackage{shortcuts}
\usepackage{booktabs}
\usepackage[capitalise]{cleveref}
\usepackage{suffix}
\usepackage{enumitem}
\usepackage{bm}

\newtheorem{lemma}{Lemma}[section]

\newcommand{\A}{\mathcal A}
\newcommand{\sumA}{\sum_{a\in\A}}
\newcommand{\sumn}{\sum_{i=1}^n}
\newcommand{\eavg}{\frac1n\sumn}
\WithSuffix\newcommand\sumA*[1]{\sum_{#1\in\A}}
\newcommand{\X}{\mathcal X}
\newcommand{\DeltaA}{\Delta^{\A}}

\DeclareMathOperator*{\esssup}{ess\,sup}

\def\blind{0}

\def\edit{}
\def\blockedit{}
\if\blind1
\usepackage{xcolor}
\newcommand{\editb}{\textcolor{blue}}
\newcommand{\blockeditb}{\color{blue}}
\fi
\if\blind0
\def\editb{}
\def\blockeditb{}
\fi

\begin{document}

\def\spacingset#1{\renewcommand{\baselinestretch}%
{#1}\small\normalsize} \spacingset{1}

\newcommand{\mytitle}{More Efficient Policy Learning via\\Optimal Retargeting}
  \title{\bf \mytitle}
  \if1\blind\author{}\fi
  \if0\blind\author{Nathan Kallus\thanks{School of Operations Research and Information Engineering and Cornell Tech, Cornell University}}\fi
  \date{}
  \maketitle

\bigskip
\begin{abstract}
Policy learning can be used to extract individualized treatment regimes from observational data in healthcare, civics, e-commerce, and beyond. One big hurdle to policy learning is a commonplace lack of overlap in the data for different actions, which can lead to unwieldy policy evaluation and poorly performing learned policies. We study a solution to this problem based on retargeting, that is, changing the population on which policies are optimized. We first argue that at the population level, retargeting may induce little to no bias. We then characterize the optimal reference policy and retargeting weights in both binary-action and multi-action settings. We do this in terms of the asymptotic efficient estimation variance of the new learning objective. 
Extensive empirical results in a simulation study and a case study of personalized job counseling demonstrate that retargeting is a fairly easy way to significantly improve any policy learning procedure applied to observational data.
\end{abstract}

\noindent%
{\it Keywords:} 
individualized treatment regimes,
overlap,
efficient policy learning,
optimization.
\spacingset{1.5} %

\section{Introduction}\label{sec:intro}

Personalized intervention policies, also known as individualized treatment regimes, are having increasing impact in domains such as education \citep{mandel2014offline}, healthcare \citep{bertsimas2017personalized}, and public policy \citep{kube2019allocating}.
In many of these domains exploration is costly, unethical, or otherwise prohibitive, and so it is crucial to learn new policies using existing observational data. The potential for this can be big: in medicine, for example, the widespread adoption of electronic medical records provides very rich and plentiful data that can help drive the future of personalized care \citep{kosorok2019precision}.

Learning new policies from observational data relies on our ability to evaluate a counterfactual: on average, how would an individual have fared under a treatment different than that given in the past. 
If we have rich enough covariates we may be able to defend an assumption of unconfounded treatment assignment, in which case the solution is essentially to compare like to like: on average the individual would fare like a similar individual who received the alternative treatment. For this to be possible, of course, a like comparison must be available in the data. The lack of sufficiently similar comparisons is known as \emph{limited overlap} and it can lead to highly unwieldy policy value estimates and poorly performing learned policies in practice.

In this paper we study one way to address this problem: change the population on which we optimize the policy, which we term \emph{retargeting}. We demonstrate that in many cases the optimal policy can actually be characterized as the population minimizer of a variety of population objective functions, so it stands to reason that we should use the one that is easiest to estimate. Each objective is centered with respect to a different reference policy and averaged over a different weighted subpopulation. 
We characterize the estimability of these objectives in terms of their efficiency bounds, that is, the minimal asymptotic variance that a consistent regular estimator can have in estimating them.
We show how to optimize these bounds in the binary- and multiple-action cases. 
We demonstrate these new tools empirically in an extensive simulation study and in an application to personalized job counseling. 
The results show that retargeting can offer significant benefits and is fairly robust. In sum, retargeting is an easy way to improve any policy learning procedure using observational data where limited overlap is a major concern.

\subsection{Problem set up}\label{sec:setup}

A policy $\pi$ is a map from the space $\X$ of baseline covariates to a probability distribution over the action space $\A$. Specifically, let $\Delta^{\A}=\braces{p\in\R{\A}_+:\sumA p(a)=1}$ be the space of probability distributions over actions and let $\Pi=\prns{\Delta^{\A}}^\X$ be the space of all maps $\X\to\Delta^{\A}$. We index an element $\pi\in\Pi$ by letting $\pi(a\mid x)$ be the $a$ coordinate in the image of $x$ under the map $\pi$, \ie, the probability of taking action $a$ when seeing baseline covariates $x$.

The population of decision instances is described by covariates $X\in\X$ and potential outcomes $(Y(a))_{a\in\A}\in\R{\A}$. That is, in each decision instance, first the covariates $X$ are revealed, then we take some action $a\in\A$, and then we get the reward $Y(a)$. Let $\mu(a\mid X)=\Eb{Y(a)\mid X}$ and $\sigma^2(a\mid X)=\op{Var}(Y(a)\mid X)$.
The value of a policy $\pi$ is the average reward we get over the population of instances when we choose our action from the distribution given by $\pi(\cdot\mid X)$:
$$
V(\pi) = \Eb{\sumA\pi(a\mid X)Y(a)}=\Eb{\sumA\pi(a\mid X)\mu(a\mid X)}.
$$
Note that the last expectation is \emph{only} over the variable $X$, where the inside of the expectation is the expected reward within the context of seeing an instance with covariates $X$.
The best possible policy value is $V^*=\max_{\pi\in\Pi}V(\pi)$.

The data available for learning policies consists of $n$ observations of covariates, action, and reward, $X_1,A_1,Y_1,\dots,X_n,A_n,Y_n$. We drop subscripts to indicate a generic draw. We assume that the observed reward is the potential reward of the observed action, $Y=Y(A)$, encapsulating non-interference and consistency, also known as stable unit value assumption \citep[SUTVA;][]{rubin1980randomization}. We further assume that the data is mean-unconfounded in that $\mu(a\mid X)=\Eb{Y\mid A=a,X}\,\forall a\in\A$. We let $\phi(a\mid X)=\Prb{A=a\mid X}$ be the propensity score.

\subsection{Related Literature}\label{sec:relatedlit}

\paragraph{Policy learning.}
Since an optimal policy is given by $\pi(a^*(X)\mid X)=1$ for any $a^*(X)\in\argmax_{a\in\A}\mu(a\mid X)$, one approach to policy learning is to fit a regression estimate $\hat\mu$ and plug it into $a^*$ above.
Under unconfoundedness, $\hat\mu$ can be estimated by regressing the observed outcome $Y$ on $A,X$.
This policy is called \emph{direct comparison} (DC) and in the absence of additional assumption or constraints it enjoys certain minimax guarantees \citep{hirano2009asymptotics,stoye2009minimax}.
However, often we are interested in a more structured policy, but at the same time have to deal with confounding. For example, we might want the policy to not consider ethnoracial features but we do need to consider them in accounting for confounding factors. Additionally, we may want the policy to be simple and interpretable, like a separating hyperplane or a decision tree.
In such cases, we want to restrict ourselves to a constrained policy class $\Pi_0\subseteq\Pi$.
This may also afford us additional efficiency as we are only truly interested in a potentially much lower dimensional object than $\mu$ itself.

In such cases, various methods have been proposed, which estimate $V(\pi)$ and then optimize this estimate over $\pi\in\Pi_0$ (we embed any potential regularization implicitly as constraints on $\Pi_0$ itself). 
Estimates of $V(\pi)$ are similar to estimates of averages with missing-at-random data or of average treatment effects under unconfoundedness.
One policy learning approach is the \emph{direct method} (DM), which uses the regression-adjusted estimate $\hat V_n(\pi)=\eavg\sumA\pi(a\mid X_i)\hat\mu(a\mid X_i)$ given a fitted $\hat\mu$ \citep{qian2011performance}.
Another is to use \emph{inverse propensity weighting} (IPW), which, given a fitted $\hat\phi$, estimates $\hat V_n(\pi)=\eavg\pi(A_i\mid X_i)Y_i/\hat\phi(A_i\mid X_i)$ \citep{beygelzimer2009offset,li2011unbiased}.
When $\Pi_0$ has bounded capacity, bounds on the regret of the learned policy compared to the a-priori-best policy in $\Pi_0$ can be obtained via uniform concentration inequalities on $\hat V_n(\pi)$ \citep{kitagawa2018should}.
Optimizing such estimates $\hat V_n(\pi)$ is a form of cost-sensitive classification and can be approached using convex surrogate losses \citep{zhao2012estimating}.
In the absence of unconfoundedness, $V(\pi)$ may be unidentifiable but well-performing policies may still be learned \citep{kallus2018confounding}.

To deal with the instability of inverse probability weights, normalization (H\'ajek estimates) and clipping are often employed \citep{swaminathan2015self}. Balanced policy learning \citep{kallus2018balanced} uses optimal balance \citep{kallus2016generalized} to deal more directly with this instability.
Doubly robust (DR) policy learning \citep{dudik2011doubly} uses the augmented IPW (AIPW) estimator \citep{robins1994estimation}. Using cross-fitted models makes this estimate efficient under mild conditions \citep{chernozhukov2017double}, which has been shown to lead to better regret bounds \citep{athey2017efficient}.
IPW and DR can also be generalized to continuous action spaces \citep{chen2016personalized,kallus2018policy}.

We further discuss the statistical efficiency of various policy value estimates and its relationship to policy learning in \cref{sec:efficientestimation}.

\paragraph{Dealing with limited overlap in average treatment effect estimation.}
Overlap refers to the extent to which the covariate distributions conditioned on different observed actions overlap with one another in the data, which controls the availability of like comparisons.
As richer covariates are used to better support the plausibility of unconfoundedness and to enable more pinpoint personalization of policies, the issue of overlap becomes more dire \citep{d2017overlap}.
Overlap is often quantified by how close to 0 or 1 the propensities $\phi(a\mid X)$ are at any point $X$. Lack of overlap leads to a fundamental difficulty in estimating causal estimands \citep{lalonde1986evaluating,heckman1997matching,dehejia1999causal}.
In the estimation of $\hat\mu$ it can be understood as a covariate shift between the distribution where $\hat\mu(\cdot\mid a)$ is fit and where it is deployed.
This shift means that parametric models will precariously extrapolate from areas of high density to areas of low density, while non-parametric models will have little local information to drive predictions. In IPW and AIPW estimates, the size of the density ratio $\pi(A\mid X)/\phi(A\mid X)$ drives the variance and grows big as overlap diminishes.

In the context of estimating average treatment effects, \ie, $\Eb{Y(+)-Y(-)}$ or\break $\Eb{Y(+)-Y(-)\mid A=+}$ with $\A=\{-,+\}$, various approaches to dealing with this fundamental issue have been investigated. Clipping case weights \citep{ionides2008truncated} or regularizing them \citep{kallus2016generalized,santacatterina2018optimal} shrinks the estimate toward the unadjusted one to trade bias for variance. In the context of matching, another approach is to drop units without good matches \citep{iacus2011multivariate,rubin2010limitations,cochran1973controlling}. Similarly, one can drop units with extreme propensities \citep{dehejia1999causal,heckman1997matching,smith2005does}. Both limit the estimates to a subpopulation with better overlap. \citet{crump2009dealing} formalize this notion and frame it as choosing a subpopulation so to optimize (part of) the efficiency bound of the corresponding estimand. They show that thresholding the propensity score gives the best subpopulation defined as a subset of $\X$ and they also characterize the best weighted subpopulation, which under homoskedasticity is reweighted by $\phi(+\mid X)\phi(-\mid X)$. \citet{li2018balancing} consider these latter weights and show that they correspond to certain balancing conditions.
All of these lead to retargeted estimated effects that are local to a subpopulation different from the study one.

In the context of policy learning, study of the ramifications of limited overlap and ways to address it has been limited.
\citet{swaminathan2015counterfactual} regularize to force the learned policy to be close to $\phi$ in order to control the size of density ratios. Moreover, weight clipping \citep{swaminathan2015counterfactual} and regularization \citep{kallus2018balanced} are generally also used in policy learning. However, the modification of the target population on which policies are optimized has not been previously studied and neither has its optimal choice. We will here show it holds particularly unique promise for policy learning, which may incur little to no asymptotic bias due to retargeting, where a well performing policy is much more important than an unbiased or interpretable population estimate, and where some train-test shift is often inevitable anyway.

\section{Centering and Retargeting the Policy Learning Objective}

In order to alleviate the issues of lack of overlap we will consider \emph{changing} the objective that we optimize. Specifically, we will consider reweighting it. For every $w\in\R{\X}_{++}$, define the $w$-weighted population policy value as
$$V(\pi;w)=\Eb{w(X)\sumA\pi(a\mid X)\mu(a\mid X)}.$$
This retargets policy learning to the subpopulation weighted by $w$.
We will seek the weights $w$ for which $V(\pi;w)$ is in some sense most convenient from the point of view of estimation. 
Because the scale of our objective function does not matter, we will restrict our attention to $w$ satisfying $\Eb{w(X)}=1$.

But, in fact, the optimizer over $\pi$ of $V(\pi;w)-V(\rho;w)$ remains the same over choice of a fixed reference policy $\rho$, and the same is true of $\hat V_n(\pi;w)-\hat V_n(\rho;w)$ for any value estimator. Therefore, it is actually not immediately clear the estimability of which objective we should be considering. We therefore will \emph{also} seek the best reference policy $\rho$ that makes the centered value the easiest to estimate. Even though centering by a reference policy has no impact on the policy learning, finding the best reference policy allows us to correctly characterize the estimability of different retargeted learning objectives.

Specifically, for every $w\in\R{\X}_{++}$ and $\rho\in\R{\A\times\X}$, define 
$$R(\pi;w,\rho)=\Eb{w(X)\sumA(\pi(a\mid X)-\rho(a\mid X))\mu(a\mid X)}.
$$
We interpret this as the additional average value of $\pi$ compared to $\rho$ over the retargeted population where contexts are reweighted by $w(X)$.
Letting $\bm 1(X)=1$ we recognize $R(\pi;\bm 1,\rho)=V(\pi)-V(\rho)$.
Note that we do not actually constrain $\rho$ to be a policy, that is, its image need not be nonnegative and sum to one. 
We therefore also recognize $V(\pi;w)=R(\pi;w,\bm 0)$.

In the next sections we will first consider how this new objective impacts the learning problem and then discuss its efficient estimation from data for a fixed $w,\rho$.

\subsection{The Effect of Retargeting on Policy Learning}\label{sec:invariance}

In policy learning, we seek a policy that optimizes $V(\pi)$. Define the corresponding set of optimal policies as
\begin{equation}\Pi^*=\argmax_{\pi\in\Pi} V(\pi).\label{eq:oldopt}\end{equation}
Note that $\pi^*\in\Pi^*$ if and only if $\pi^*(a\mid X)(\max_{a'\in\A}\mu(a'\mid X)-\mu(a\mid X))=0\,\forall a\in\A$ almost surely. That is, $\pi^*(\cdot\mid X)$ can only support actions with largest conditional mean reward given $X$.

As discussed in \cref{sec:relatedlit}, in practice we only optimize over a restricted policy class, $\Pi_0\subseteq\Pi$. Let the best policies in $\Pi_0$ be denoted by
\begin{equation}\Pi_0^*=\argmax_{\pi\in\Pi_0} V(\pi).\label{eq:oldopt0}\end{equation}

We next consider the question of what happens if we instead optimize our alternative objective $R(\pi;w,\rho)$ over $\Pi_0$. 
Let
\begin{equation}
 \Pi^*_0(w,\rho)=\argmax_{\pi\in\Pi_0} R(\pi;w,\rho).\label{eq:newopt}
\end{equation}

First, notice that $R(\pi;w,\rho)$ decomposes as a sum of the weighted value of $\pi$ minus the weighted value of $\rho$. Therefore, as also discussed above, the choice of $\rho$ has no influence on the optimizer of $R(\pi;w,\rho)$ no matter what set of policies one optimizes over, as it only amounts to constant shift in the objective. That is, $ \Pi_0^*(w,\rho)= \Pi_0^*(w,\rho')$. In particular, we immediately see that $ \Pi_0^*(\bm1,\rho)=\Pi_0^*$ for any $\rho$.

Next, we consider the effect of weighting. As noted above, an optimal unconstrained policy simply selects the best action in each context. If we reweight the contexts, we will arrive at the same conclusion. But this may not be true of a constrained optimal policy. We next establish that if there exists some policy in $\Pi_0$ that is also optimal in an \emph{unconstrained} way, then neither retargeting nor centering affects policy learning over $\Pi_0$ at the population level.
\begin{lemma}\label{lemma:weightinvariance}
Suppose $\Pi^*\cap\Pi_0\neq\varnothing$.
Then $ \Pi_0^*(w,\rho)=\Pi^*\cap\Pi_0$ for every $w\in\R{\X}_{++}$, $\rho\in\R{\A\times\X}$. In particular, if $\pi$ is a solution to \cref{eq:newopt} then $V(\pi)=V^*$.
\end{lemma}

\Cref{lemma:weightinvariance} shows that if $\Pi_0$ is well-specified, in that it contains one of the true unconstrained optimal policies, then policy learning over the class $\Pi_0$ is invariant to retargeting.
In other words, there are many objectives that characterize the best policy in $\Pi_0$.
\editb{This, of course, depends on well-specification; we discuss the case of misspecification in \cref{sec:discussion misspecification and shift}.} 

{\blockedit
A result akin to \cref{lemma:weightinvariance} also applies in a given sample.
Let \begin{align*}
V_n(\pi)&=\frac1n\sum_{i=1}^n\sumA\pi(a\mid X_i)\mu(a\mid X_i),\\
R_n(\pi;w,\rho)&=\frac1n\sumn w(X_i)\sumA(\pi(a\mid X_i)-\rho(a\mid X_i))\mu(a\mid X_i).
\end{align*}
respectively be the value of $\pi$ on the given sample and the corresponding sample version of $R(\pi;w,\rho)$. Similarly, let
$$
\Pi^*_n=\argmax_{\pi\in\Pi}V_n(\pi),\quad \Pi^*_{0,n}(w,\rho)=\argmax_{\pi\in\Pi_0} R_n(\pi;w,\rho).
$$
\begin{lemma}\label{lemma:weightinvariance2}
Suppose $\Pi^*\cap\Pi_0\neq\varnothing$.
Then $ \Pi_{0,n}^*(w,\rho)=\Pi^*_n\cap\Pi_0$ for every $w\in\R{\X}_{++}$, $\rho\in\R{\A\times\X}$ almost surely.
\end{lemma}

Note that $\Pi^*\subseteq \Pi^*_n$ almost surely.
The conclusion from \cref{lemma:weightinvariance2} is that if $\Pi_0$ is well-specified then optimizing the oracle sample objective, which uses perfect counterfactual predictions to score policies on the given $X$-sample, is equivalent to optimizing its retargeted version.
}

\subsection{Efficient Estimation of Retargeted Policy Value}\label{sec:efficientestimation}

{\blockedit
In the above we introduced a new estimand, $R(\pi;w,\rho)$\edit{, and its sample version, $R_n(\pi;w,\rho)$}. For it to serve as an objective for policy learning, we also need to estimate it. 
To characterize how easily estimable \edit{these are}, we use consider the corresponding asymptotic efficiency bound, that is, we focus on the asymptotic estimation variance of any efficient regular 
estimator for $R(\pi;w,\rho)$.

\begin{lemma}\label{lemma:efficientinfluence}
The efficient influence function for $R(\pi;w,\rho)$ is $\psi(x,a,y;\phi,\mu)-R(\pi;w,\rho)$ where
\begin{align*}
\psi(x,a,y;\phi_0,\mu_0)=w(x)\Biggl(&\sumA*{a'}(\pi(a'\mid x)-\rho(a'\mid x))\mu_0(a'\mid x)\\&\qquad\qquad\qquad+\frac{\pi(a\mid x)-\rho(a\mid x)}{\phi_0(a\mid x)}(y-\mu_0(a\mid x))\Biggr).
\end{align*}
Hence every efficient regular estimator for $R(\pi;w,\rho)$ must have the form
\begin{align}\label{eq:efficientform}
\hat R_n(\pi;w,\rho)=
\tilde R_n(\pi;w,\rho)+o_p(1/\sqrt{n}),~\text{where}~
\tilde R_n(\pi;w,\rho)=\frac1n\sumn\psi(X_i,A_i,Y_i;\phi,\mu).
\end{align}
Moreover,
\begin{align}
\notag n\op{Var}\Eb{\tilde R_n(\pi;w,\rho)\mid X_{1:n}}
&=n\op{Var}(\edit{R_n}(\pi;w,\rho))\\
&=\op{Var}\prns{w(X)\sumA(\pi(a\mid X)-\rho(a\mid X))\mu(a\mid X)},
\label{eq:decomposeefficientvariance1}
\\
\notag n\E\op{Var}\prns{\tilde R_n(\pi;w,\rho)\mid X_{1:n}}
&=n\Eb{\prns{\tilde R_n(\pi;w,\rho)-\edit{R_n}(\pi;w,\rho)}^2}
\\&=\Eb{
w^2(X)\sumA\frac{\sigma^2(a\mid X)}{\phi(a\mid X)}\prns{\pi(a\mid X)-\rho(a\mid X)}^2
},\label{eq:decomposeefficientvariance2}
\end{align}
And the sum of the above is the efficiency bound for $R(\pi;w,\rho)$.
\end{lemma}}

This result follows easily following the same proof arguments as in \citet{hahn1998role,hirano2003efficient}, using the results of \citet{bickel1993efficient}\edit{, and finally by the law of total variance}.

With fixed known $w,\rho$, the efficient estimation of $R(\pi;w,\rho)$ is very similar to the efficient estimation of an average from missing-at-random data. Therefore,
there are also various immediate examples of efficient estimators for $R(\pi;w,\rho)$ arising from existing literature. 
When $\mu$ is sufficiently smooth, 
the sample average of $\psi(X_i,A_i,Y_i;\bm \infty,\hat\mu)$, that is, the direct estimator, with estimated outcome regression $\hat\mu$ is efficient \citep{hahn1998role}.
When $\phi$ is sufficiently smooth, the sample average of $\psi(X_i,A_i,Y_i;\hat\phi,\bm 0)$, that is, the IPW estimator, with estimated propensities $\hat\phi$ is efficient \citep{hirano2003efficient}.
Alternatively, with less stringent smoothness assumptions, the sample average of $\psi(X_i,A_i,Y_i;\hat\phi^{(-i)},\hat\mu^{(-i)})$ is efficient when $\hat\phi^{(-i)}$, $\hat\mu^{(-i)}$ are ``cross-fold'' estimated with convergence rates $o_p(n^{-1/4})$, that is the double machine learning (DML) estimator \citep{chernozhukov2017double}.

Since all of the above efficient estimators rely on plugging in nuisance estimates of $\mu,\phi$ that are shared across policies $\pi$, it easily follows that these estimates also obtain uniform efficiency across $\pi\in\Pi_0$ (\ie, the $o_p(1/\sqrt{n})$ is uniform in $\pi\in\Pi_0$ in \cref{eq:efficientform}) as long as $\Pi_0$ has bounded capacity (\ie, has a polynomial covering number in inverse radius, which is implied by, \eg, having finite pseudo- or Vapnik-Chervonenkis dimension, \citealp{vapnik2013nature}). This follows from a straightforward invocation of standard empirical process theory arguments (see, \eg, \citealp{pollard1990empirical}); we omit the details. 
For example, (without any retargeting) \citet{athey2017efficient} use a careful application of this to obtain tighter regret bounds than are possible with optimizing inefficient policy value estimates.

\edit{\Cref{lemma:efficientinfluence}} decomposes the efficiency bound into two terms. The first\edit{, \cref{eq:decomposeefficientvariance1},} reflects only the variance due to $X$-variation in the population and depends only on $\mu$ and not $\phi$. The second\edit{, \cref{eq:decomposeefficientvariance2},} reflects the variance at any one $X$-level due to residual noise and, importantly, any lack of overlap, averaged over $X$ values, and depends only on $\phi$ and not $\mu$. {\blockedit This latter term can be understood as the ``efficiency bound'' of estimating $R_n(\pi;w,\rho)$ in the sense that if $\hat R_n$ is an efficient regular estimator then we must have
\begin{equation}\label{eq:asympnormRn}
\sqrt{n}(\hat R_n-R_n(\pi;w,\rho))\annot{d}{\longrightarrow}\mathcal N\prns{0,\;\textstyle\Eb{
w^2(X)\sumA\frac{\sigma^2(a\mid X)}{\phi(a\mid X)}\prns{\pi(a\mid X)-\rho(a\mid X)}^2}}.
\end{equation}%
}

\section{Optimal Retargeting}\label{sec:optretarget}

In the previous section, we first argued that there may be many objectives that characterize the optimal policy as their maximizer\edit{, both in the sample and in the population,} and we then presented results on how efficiently estimable these objectives are. It therefore stands to reason we should pick the objective that is most convenient, or \emph{easiest to estimate}.
We next consider precisely how to do this.

First, we discuss precisely what we will be mean by easiest to estimate. \edit{
Ideally, if we had access to prefect counterfactual predictions for all actions, we would optimize the sample policy value, $V_n(\pi)$, and not suffer from lack of overlap.
In view of \cref{lemma:weightinvariance2}, under correct specification, this is equivalent to optimizing ${R_n}(\pi;w,\rho)$ for any $w$.
Moreover, in view of \cref{lemma:weightinvariance2}, the choice of $w$ does not affect this latter optimization, so we need not consider how $w$ affects the variance of ${R_n}(\pi;w,\rho)$ itself, only how $w$ affects the variance of \emph{estimating} ${R_n}(\pi;w,\rho)$.
In other words, the asymptotic variance of any asymptotically unbiased estimator (\eg, an efficient an estimator) can be decomposed into the sum of the variance of estimating ${R_n}(\pi;w,\rho)$ and the variance of ${R_n}(\pi;w,\rho)$ itself, and since $w$ does not change the optimizer of ${R_n}(\pi;w,\rho)$ (only the optimizer of our estimate thereof), we only consider the variance of estimating ${R_n}(\pi;w,\rho)$.}

\edit{The efficiency bounds from the last section exactly characterize how difficult this estimation is.
In particular, any efficient regular estimator will have an asymptotic mean-squared error given by \cref{eq:decomposeefficientvariance2} in estimating ${R_n}(\pi;w,\rho)$.
We therefore seek the objective ${R_n}(\pi;w,\rho)$ requiring minimal such asymptotic error.
(See also \cref{sec:discussion efficiency bound} for further discussion.)}

Still, our variance in estimating $\edit{R_n}(\pi;w,\rho)$ also depends on $\pi$. When doing policy learning, we need to estimate our objective for various policies and we want the weights $w$ to be independent of the policy $\pi$ to avoid any bias. Therefore, our criterion for a good objective for policy learning will be whether it is easily estimable uniformly for \emph{all} policies.
Toward that end, we define,
\begin{align*}
\Omega(w,\rho)&=
\sup_{\pi\in\Pi}
\Eb{
w^2(X)\sumA\frac{\sigma^2(a\mid X)}{\phi(a\mid X)}\prns{\pi(a\mid X)-\rho(a\mid X)}^2
}.
\end{align*}
Our aim will be to seek $w,\rho$ with small $\Omega(w,\rho)$, \ie, that lead to uniformly easily estimable objective functions for policy learning.
Given such desirable $w,\rho$, our approach will then be to construct an efficient estimate $\hat R_n(\pi;w,\rho)$ for $R_n(\pi;w,\rho)$ and to minimize it over $\pi\in\Pi_0$.
We recall that $\rho$ actually has no effect on the policy learning, in the sample or in the population. Nonetheless, the best $\rho$ provides the best centered objective and will facilitate choosing the best retargeting weights $w$.

{\blockedit
In particular, if $\Pi_0$ is of bounded capacity, then $\Omega(w,\rho_0)$ controls the uniform deviation between our estimated objective and the ideal finite-sample objective, $V_n(\pi;w)$.
Focusing on deterministic policies ($\Pi_0\subseteq\{0,1\}^{\A\times\X}$), we characterize the capacity of $\Pi_0$ in terms of its uniform entropy integral $\kappa(\Pi_0)=\int_0^1\sup_{m,x_1,\dots,x_m}\sqrt{\log(N(\epsilon^2,\Pi_0,\{x_1,\dots,x_m\}))}d\epsilon$, where $N(\epsilon,\Pi_0,\{x_1,\dots,x_m\})$ 
is the smallest number $N$ of policies $\pi_1,\dots,\pi_N\in\Pi_0$ such that $\sup_{\pi\in\Pi_0}\inf_{j=1,\dots,N}\frac1m\sum_{i=1}^m\sumA\abs{\pi(a\mid x_i)-\pi(a\mid x_i)}\leq\epsilon$.

We first focus on $\tilde V_n(\pi;w):=\tilde R_n(\pi;w,\bm 0)$ as in \cref{eq:efficientform}.
\begin{lemma}\label{lemma:unifconv}
Suppose that $\phi(A\mid X)$ is bounded away from zero and $Y$ is bounded. Let $w\in\R{\X}_{++}$ with $\|w\|_{\infty}<\infty$ be given.
Then there exists a universal constant $C$ such that, for any $\delta\in(0,1/2)$ and $\rho\in\R{\A\times X}$, with probability at least $1-\delta$,
\begin{align*}
&\sup_{\pi,\pi'\in\Pi_0}\abs{(\tilde V_n(\pi;w)-\tilde V_n(\pi';w))-(V_n(\pi;w)-V_n(\pi';w))}
\\&\qquad\qquad\leq
C\prns{\kappa(\Pi_0)+1+\sqrt{\log(1/\delta)}}
\sqrt{\frac{\Omega(w,\rho)}n}+o\prns{\frac{\log(1/\delta)}{\sqrt{n}}}
\end{align*}
\end{lemma}
Recall that an efficient estimator must have $\abs{\hat V_n(\pi;w)-\tilde V_n(\pi;w)}=o_p(1/\sqrt{n})$. If $\Pi_0$ is of bounded capacity, then we can usually obtain \emph{uniformly} efficient estimators, $\sup_{\pi\in\Pi_0}\abs{\hat V_n(\pi;w)-\tilde V_n(\pi;w)}=o_p(1/\sqrt{n})$ (see, \eg, \citealp[Lemma 3]{zhou2018offline}), so that \cref{lemma:unifconv} similarly applies to $\hat V_n(\pi;w,\rho)$. (See also the discussion in \cref{sec:dicussion regret,sec:dicussion local}.)

Before proceeding to discuss how to optimize our uniform efficiency objective, $\Omega(w,\rho)$, we first reformulate it in more digestible terms.}
\begin{lemma}\label{lemma:omegaform}
\begin{align*}
\Omega(w,\rho)&=
\Eb{
w^2(X)
\prns{
\sumA\frac{\sigma^2(a\mid X)}{\phi(a\mid X)}\rho^2(a\mid X)
+\max_{a\in\A}\frac{\sigma^2(a\mid X)}{\phi(a\mid X)}(1-2\rho(a\mid X))
}
}.
\end{align*}
\end{lemma}

\subsection{The Binary-Action Case}\label{sec:twoarmcase}

We first the consider the optimal choice of $w,\rho$ in the special case with just two actions, 
$\A=\{-,+\}$. We will generalize this to multiple actions in the next section, but we study this simple case first in separation because of the simple, elegant solution that arises and the connections to other work.

When $\A=\{-,+\}$, 
for $\pi\in\Pi$ we use the notation $\pi(x)=\pi(+\mid x)-\pi(-\mid x)=\break2\pi(+\mid x)-1$.
In particular, deterministic policies $\pi$ have $\pi(x)\in\{-1,+1\}$. 
We similarly apply this notation to the propensity score $\phi$ (which is also in $\Pi$). In particular, note that ${1\pm\phi(x)}=2\phi(\pm\mid x)$ and that $4\phi(+\mid x)\phi(-\mid x)=1-\phi^2(x)$.

\begin{lemma}\label{lemma:twoarmomega}
Let $\rho_0(+\mid x)=\rho_0(-\mid x)=\frac12$ and 
$w_0(x)\propto\prns{\frac{\sigma^2(+\mid x)}{1+\phi(x)}+\frac{\sigma^2(-\mid x)}{1-\phi(x)}}^{-1}$ such that $\Eb{w_0(X)}=1$. Then,
\begin{align*}
\rho_0&\in\argmin_{\rho}\Omega(w,\rho)\ \forall w,\\
(\rho_0,w_0)&\in\argmin_{\rho,\,w\;:\;\Eb{w(X)}=1}\Omega(w,\rho),\\
\Omega(w_0,\rho_0)&=\frac12\Eb{\prns{\frac{\sigma^2(+\mid X)}{1+\phi(X)}+\frac{\sigma^2(-\mid X)}{1-\phi(X)}}^{-1}}^{-1},\\
R(\pi;w_0,\rho_0)&=\frac{\Eb{\prns{\frac{\sigma^2(+\mid X)}{1+\phi(X)}+\frac{\sigma^2(-\mid X)}{1-\phi(X)}}^{-1}\pi(X)(\mu(+\mid X)-\mu(-\mid X))}}{2\Eb{\prns{\frac{\sigma^2(+\mid X)}{1+\phi(X)}+\frac{\sigma^2(-\mid X)}{1-\phi(X)}}^{-1}}}.
\end{align*}
\end{lemma}

\Cref{lemma:twoarmomega} shows that the best reference policy is complete randomization. Note that we did not constrain $\rho$ to be a policy, that is, to be in $\Pi$; it is such at optimality. Additionally, surprisingly, $\rho_0$ does \emph{not} depend on $\phi$. 
Surprisingly this also coincides with existing practice:
because $V(\pi)-V(\rho_0)$ is conveniently the $\pi(X)$-weighted average treatment effect, this form of centering is already \edit{often} used in policy learning with binary actions \citep{beygelzimer2009offset,athey2017efficient}. Nonetheless, just centering by a reference policy does not actually change the policy learning optimization problem, whether in the sample or the population. For our purpose, the importance of centering is in obtaining the optimal retargeting weights, which \emph{does} change the optimization problem \editb{(but not the population learning target per \cref{lemma:weightinvariance})}.

Because the optimal reference policy is complete randomization and because centering by complete randomization means that $R(\pi;w,\rho_0)$ is in fact a $w(X)\pi(X)$-weighted average treatment effect, the optimal retargeting weights we obtain are very similar to the familiar overlap weights of \citet{crump2009dealing,li2018balancing}. 
However, as long as $\Pi_0$ remains well-specified, we are \emph{not} changing the learning target, which remains the optimal policy. Moreover, when we generalize to multiple actions, the connection to weighted average treatment effects will break, but we will still be able to obtain optimal retargeting weights.

Note that if $\sigma^2(\pm\mid x)=\sigma^2$ is almost surely constant, then the results of \cref{lemma:twoarmomega} simplify to:
\begin{align*}
w_0(x)&\propto 1-\phi^2(X),\\
\Omega(w_0,\rho_0)&=\frac{\sigma^2/2}{1-\Eb{{{\phi^2(X)}}}},\\
R(\pi;w_0,\rho_0)&=\frac1{2(1-\Eb{\phi^2(X)})}{\Eb{(1-\phi^2(X))\pi(X)(\mu(+\mid X)-\mu(-\mid X))}}.
\end{align*}
Compare this to no retargeting: $\Omega(\bm 1,\rho_0)=\frac{\sigma^2}{2}\Eb{(1-\phi^2(X))^{-1}}$. Optimality of course guarantees that $\Omega(w_0,\rho_0)\leq \Omega(\bm 1,\rho_0)$, but we can also see this now directly from Jensen's inequality, where we would have equality if and only if $\phi(X)$ were almost surely constant.

Finally, note that $w_0$ is of course not generally known. In the homoskedastic case, for example, because it depends on the unknown propensity score, the efficiency bound for $R(\pi;w_0,\rho_0)$ is \emph{different} from simply plugging in $w_0(x)\propto 1-\phi^2(X)$ into \edit{the sum of \cref{eq:decomposeefficientvariance1,eq:decomposeefficientvariance2}}. However, the additional inflation in the efficiency bound is \emph{solely} in \edit{the former term, \ie, the $X$-variation term} (and also only \emph{decreases} with less overlap). That is, for any efficient regular estimator $\hat R_n$, the asymptotic variance of $\sqrt{n}\prns{\hat R_n-R_n(\pi;w_0,\rho_0)}$ remains as in \cref{eq:decomposeefficientvariance2} with this $w_0$ plugged in. 
Since we focus solely on the this latter variance, we again ignore the inflation in the efficiency bound \edit{due to unknown $w_0$} in our consideration of optimality of retargeting weights. 

\subsection{The Multiple-Action Case}

We next treat the more general case with any number of actions, $\A=\{1,\dots,m\}$.

{\blockedit\begin{lemma}\label{lemma:multiarmomega}
Let
\begin{align*}
\xi(x)&=\prns{m-2}\prns{\sumA\frac{\phi(a\mid x)}{\sigma^2(a\mid x)}}^{-1}\\
\rho_0(a\mid x)&=\frac12\prns{1-\frac{\phi(a\mid x)}{\sigma^2(a\mid x)}\xi(x)}\\
\kappa(x)&={\sumA\frac{\sigma^2(a\mid x)}{\phi(a\mid x)}+\frac {\xi(x)}2}\\
w_0(x)&\propto\kappa^{-1}(x)\ \text{such that $\Eb{w_0(X)}=1$}.
\end{align*}
Then \begin{align*}
\rho_0&\in\argmin_{\rho}\Omega(w,\rho)\ \forall w,\\
(\rho_0,w_0)&\in\argmin_{\rho,\,w\;:\;\Eb{w(X)}=1}\Omega(w,\rho),\\
\Omega(w_0,\rho_0)&=\frac14\Eb{\kappa^{-1}(X)}^{-1},\\
R(\pi;w_0,\rho_0)&=\Eb{\kappa^{-1}(X)}^{-1}\Eb{\kappa^{-1}(X)\sumA(\pi(a\mid X)-\rho_0(a\mid X))\mu(a\mid X)}.\end{align*}
\end{lemma}

Note that if $m=2$ then $\xi(x)=0$ and we immediately recover the binary-action result of \cref{lemma:twoarmomega}, which showed that the optimal retargeting weights for policy learning coincided with the overlap weights of \citet{crump2009dealing,li2018balancing} for average effect estimation. For $m>3$, the optimal reference policy and retargeting weights we obtain above are wholly novel and do not resemble anything previously used for either policy learning or average effect estimation. 
Indeed, because there is no clear choice for a reference policy with multiple actions, as there is for binary actions, previous studies on multiple-action policy learning use no reference-policy centering \citep{kallus2018balanced,kallus2017recursive,zhou2018offline}.
Note, in particular, that unlike the binary-action case, here the optimal reference policy depends on $\phi$.
Further note that, again, we did not constrain $\rho$ to actually be a policy, that is, to be in $\Pi$, but it is such at optimality. That is, \edit{the $\rho_0(a\mid x)$ above} is nonnegative and sums to $1$ \edit{over $a\in\A$} for each $x$.

Note that if $\sigma^2(a\mid x)=\sigma^2$ is constant across $a$ and almost every $x$, then the results of \cref{lemma:multiarmomega} simplify to:
\begin{align*}
\rho_0(a\mid x)=\frac12(1-(m-2)\phi(a\mid x)),\\
w_0(x)\propto\prns{\sumA\frac1{\phi(a\mid x)}+\frac{m}2-1}^{-1}.
\end{align*}
In this case it becomes clear that the weights $w_0$ penalize any $x$ at which some propensities are small and therefore focuses on areas with good overlap, generalizing a scalarized notion of overlap to multiple treatments in a rigorous and newly relevant way.

We can also compare the efficiency bound objective to no retargeting.
It can be shown that $\Omega(\bm1,\rho_0)=\frac14\Eb{\kappa(X)}$.
Optimality guarantees $\Omega(w_0,\rho_0)\leq\Omega(\bm1,\rho_0)$, but Jensen's inequality reveals that the inequality is strict as long as $\kappa(X)$ is not almost surely constant. In the homoskedastic case, $\kappa(X)=\sigma^2\prns{\sumA\phi^{-1}(a\mid X)+(m-2)}$, which is almost surely constant if and only if $\sumA\phi^{-1}(a\mid X)$ is almost surely constant. The more it varies, the greater the benefit from retargeting.
}

\section{Empirical Results}\label{sec:emp}

We next study the added value of retargeting empirically.
We first consider a simulation study where we can vary various knobs and then a case study based on a multiarm job counseling experiment.
\edit{Replication code is available at \if0\blind\\\texttt{https://github.com/CausalML/RetargetedPolicyLearning}\fi\if1\blind[ANONYMIZED FOR REVIEW]\fi.}

\subsection{Simulation Study for Binary Actions}\label{sec:sim}

For our simulation study we focus on the following data generating process with two actions and two covariates, parameterized by $q_1,q_2,\beta\geq0$:
\begin{align*}
X''&\sim \op{Unif}[-1,1]^2,&\ X'_j&=\op{sign}(X''_j)\abs{X''_j}^{q_1},\ j=1,2,\\
Y(-)&=X'_1+\epsilon,\ \epsilon\sim\mathcal N(0,1),&\ 
Y(+)&=Y(-)+X'_1+X'_2+1/4,\\
X_j&=\op{sign}(X'_j)\abs{X'_j}^{q_2},\ j=1,2,&\ A&\sim\op{Ber}(\edit{\Phi(\beta\,X'_1)}).
\end{align*}
In the above, $\Phi$ is used to denote the cumulative distribution function of the standard normal. Only $X$ is returned as the covariate data, not $X'',X'$.
The parameter $q_1$ \edit{therefore} changes the distribution of $X$. We use it to control the level of covariate shift between training and test by using different values for the two datasets. The parameter $\beta$ controls the amount of overlap in the data. When $\beta=0$, there is perfect overlap\edit{, in the sense that at every $X$, the propensity for each action is equal}. As $\beta$ grows overlap becomes minimal\edit{, in the sense that at every $X$, the sum of inverse propensities action grows as $\exp(\Theta(\beta^2))$, using Landau notation}.
The parameter $q_2$ controls the misspecification of a linear policy. 
Because $Y(+)-Y(-)$ is linear in $X'$, when $q_2=1$ a linear policy in $X$ is well-specified. However, when $q_2\neq 1$ a linear policy in $X$ is misspecified since $X$ is a nonlinear transform of $X'$.

\begin{figure}[t!]\centering
\def\arraystretch{0}%
\setlength{\tabcolsep}{-.25em}%
\begin{tabular}{cccc}
\includegraphics[width=0.29\textwidth]{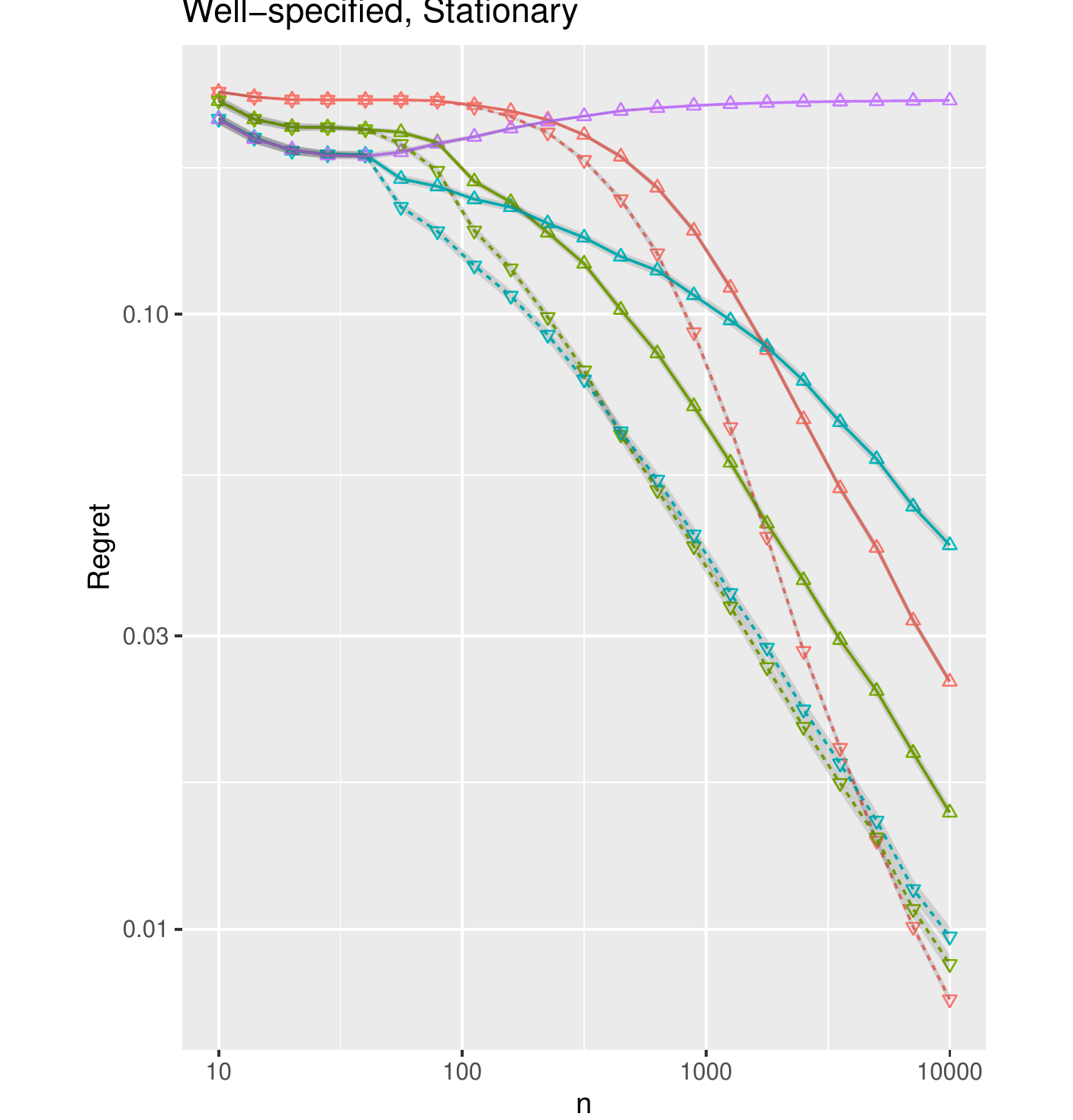}
&
\includegraphics[width=0.29\textwidth]{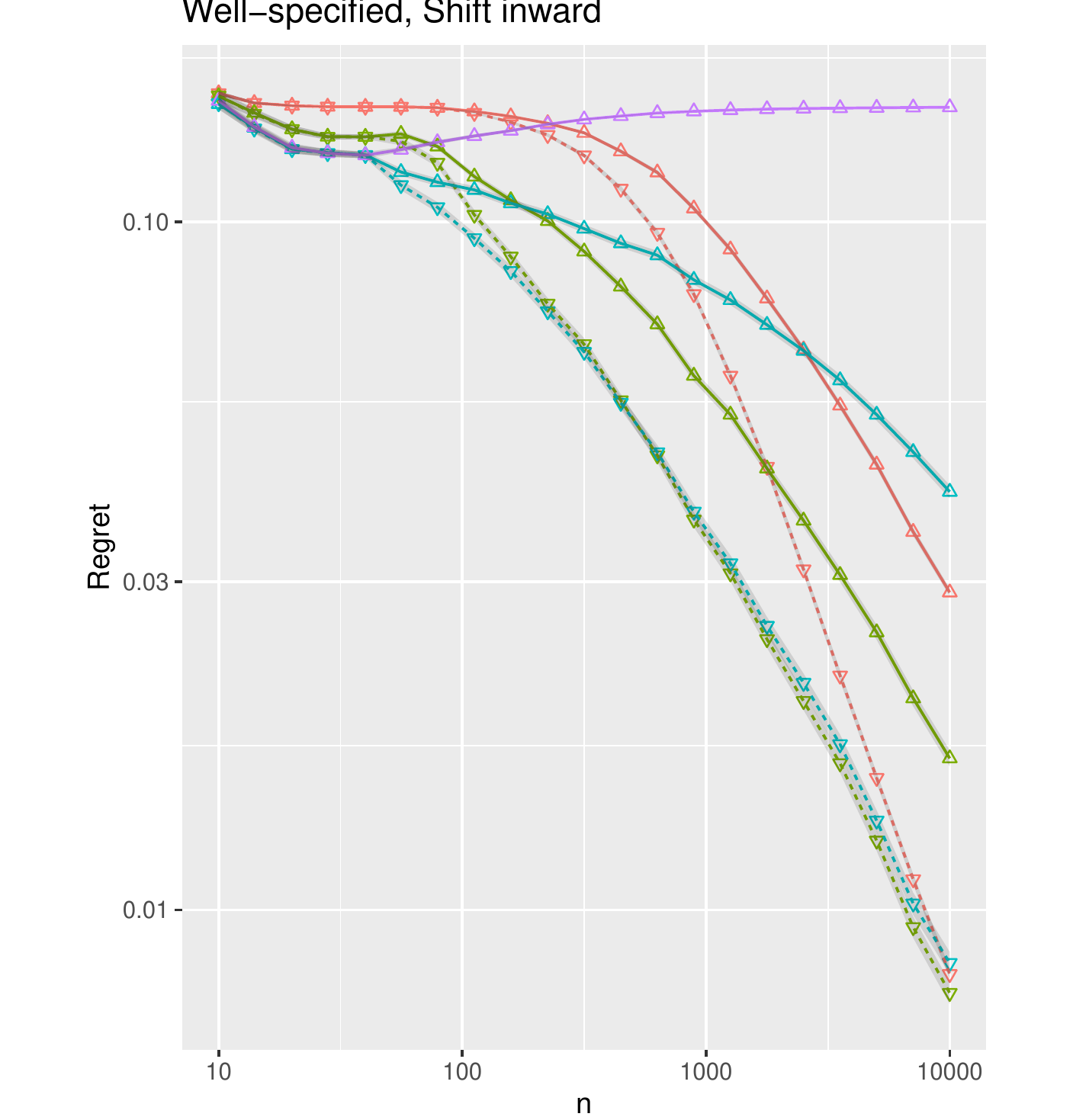}
&
\includegraphics[width=0.29\textwidth]{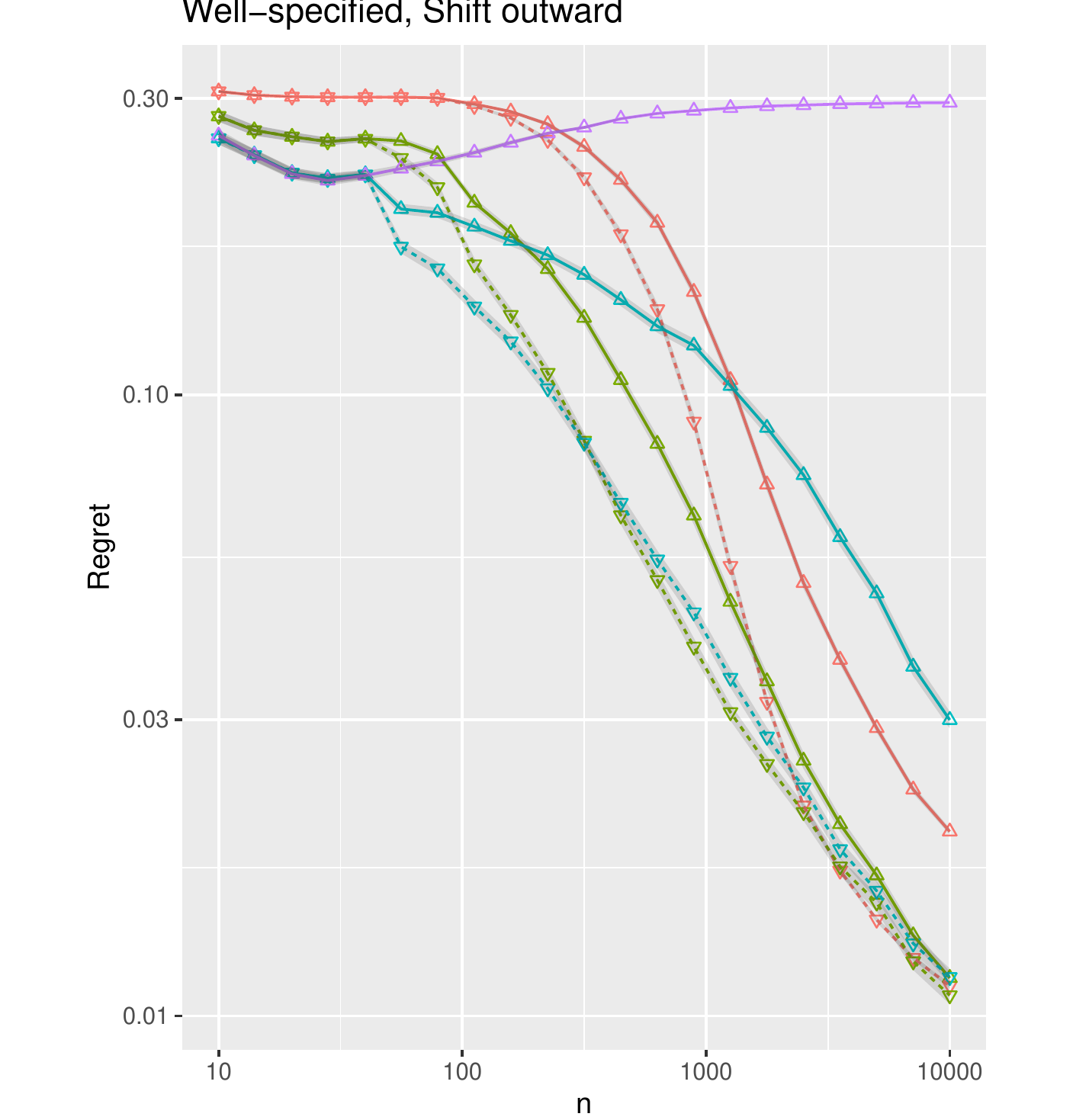}
&
\multirow{-10}{*}[6em]{\hspace{0.025\textwidth}\includegraphics[width=0.125\textwidth]{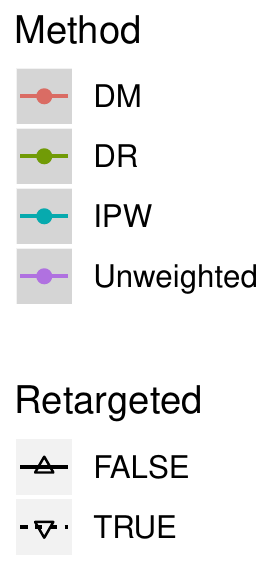}}
\\[0.5em]
\includegraphics[width=0.29\textwidth]{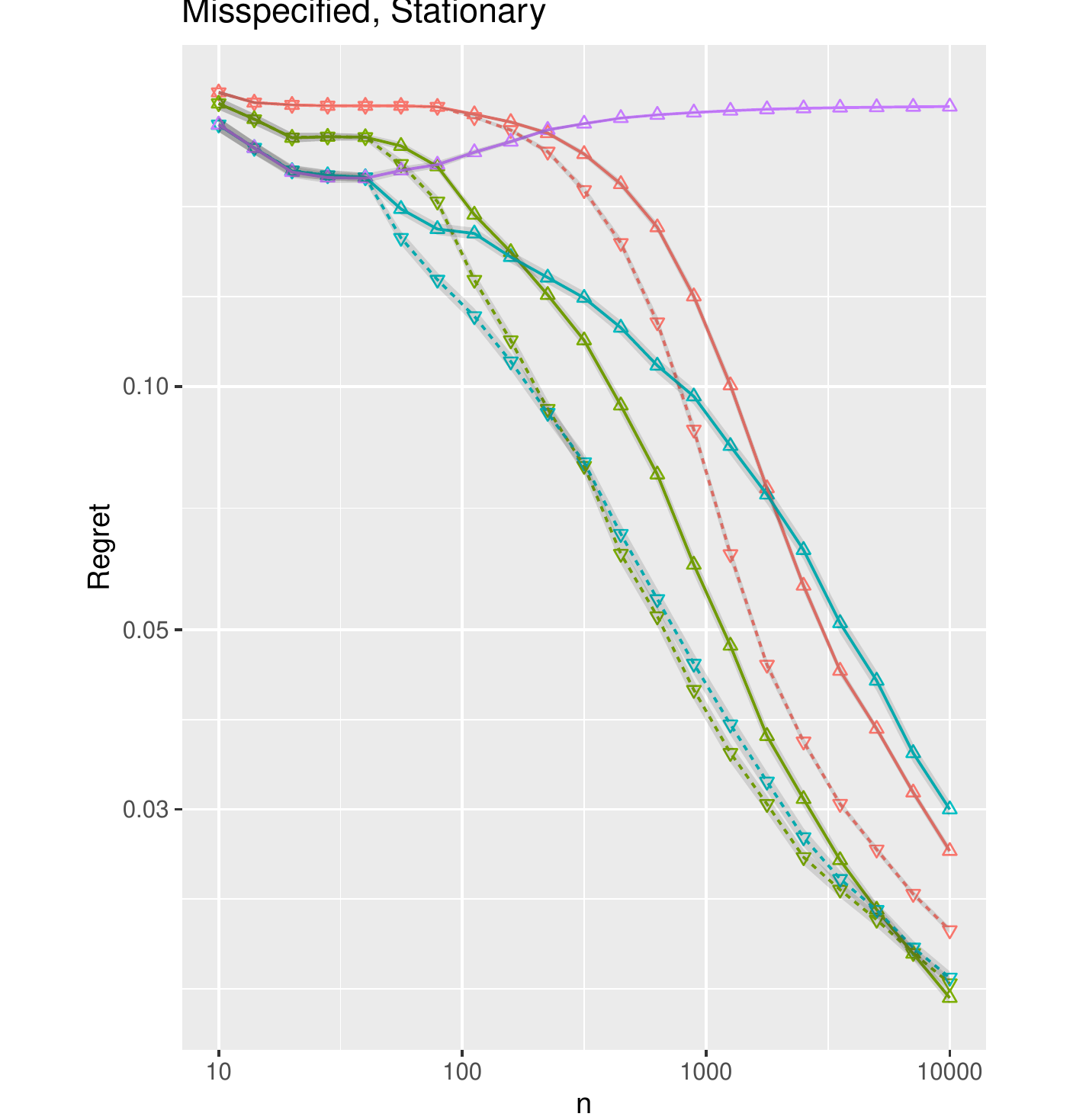}
&
\includegraphics[width=0.29\textwidth]{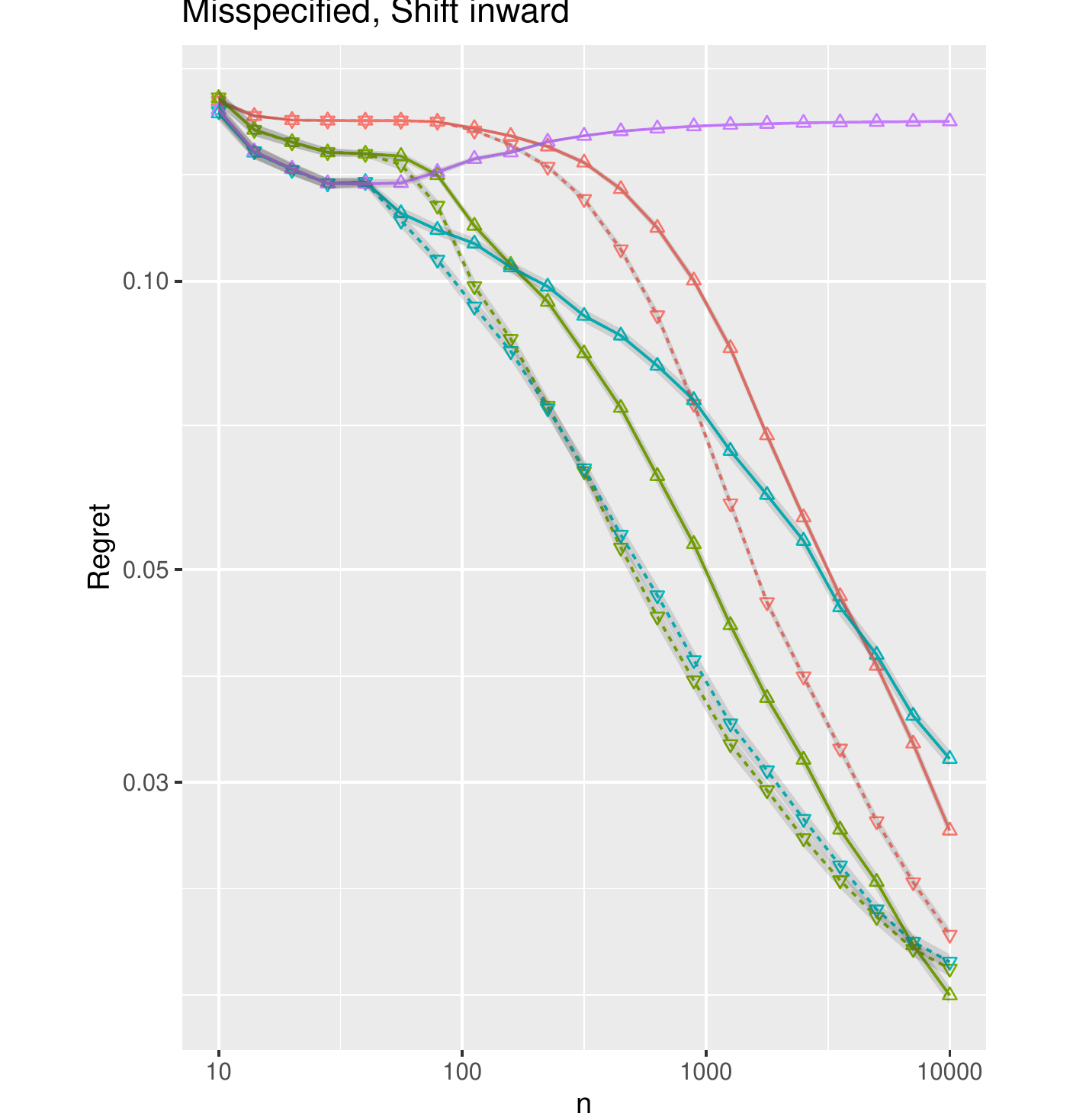}
&
\includegraphics[width=0.29\textwidth]{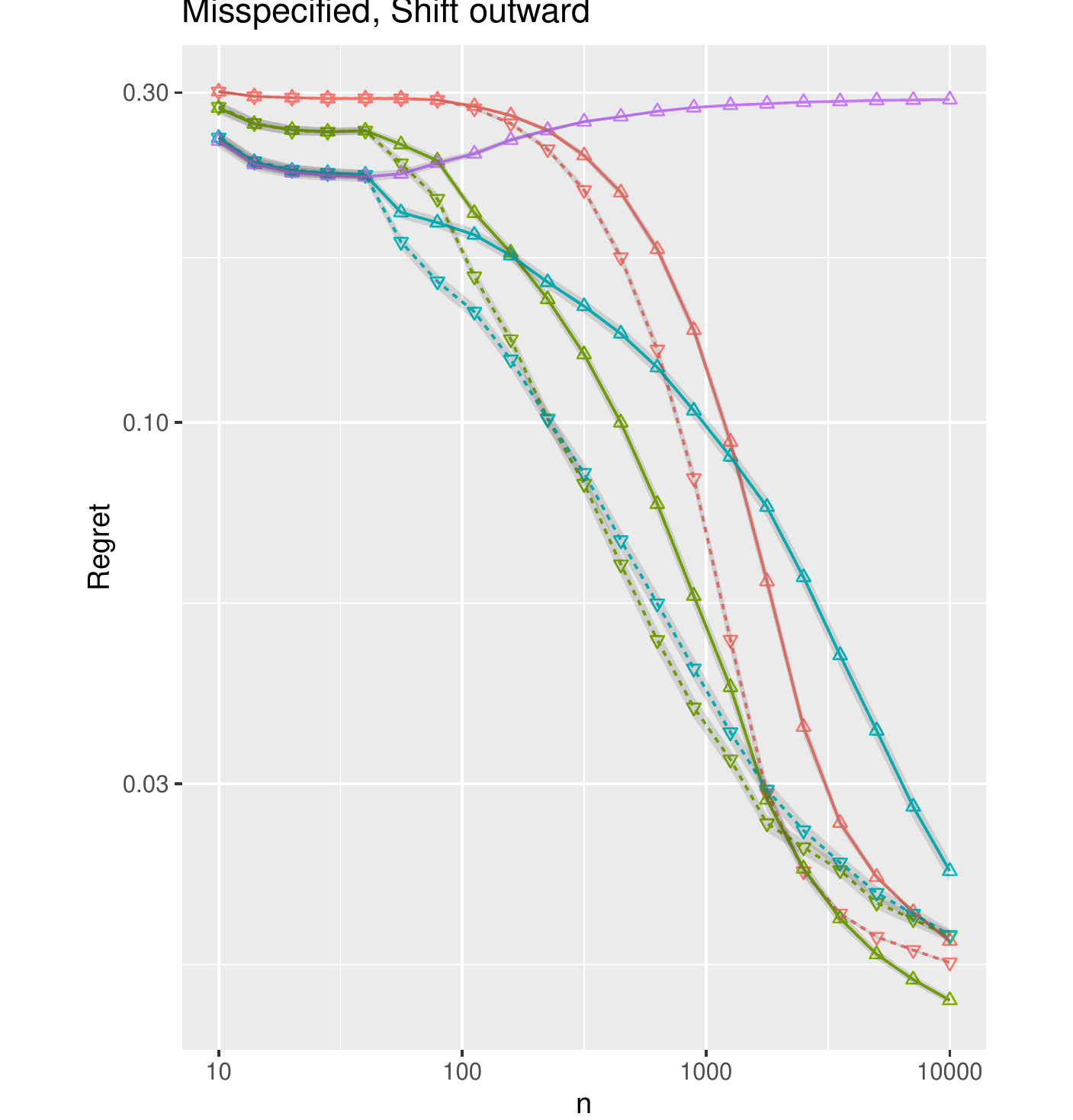}
\end{tabular}%
\caption{Average regret as training data size $n$ varies ($\beta=3.5$)}\label{fig:synthetic_n}
\end{figure}

Specifically we will consider scenarios where both training and test data are drawn with $q_2=1$ (``well-specified'') and with $q_2=1/2$ (``misspecified''). We will always take $q_1=1$ for training, but we will consider drawing the test data either with $q_1=1$ (``stationary''), $q_1=2$ (``shift inward''), or $q_1=1/2$ (``shift outward''). Out-of-sample policy values $V(\pi)$ are computed using the known $\mu$ on a test set of 100,000 draws. Regret of a policy $\pi$ is computed as $\eavg\prns{\max_{a\in\A}\mu(X_i\mid a)-\sumA\pi(a\mid X_i)\mu(X_i\mid a)}$ on the test set.

\begin{figure}[t!]\centering
\def\arraystretch{0}%
\setlength{\tabcolsep}{-.25em}%
\begin{tabular}{cccc}
\includegraphics[width=0.29\textwidth]{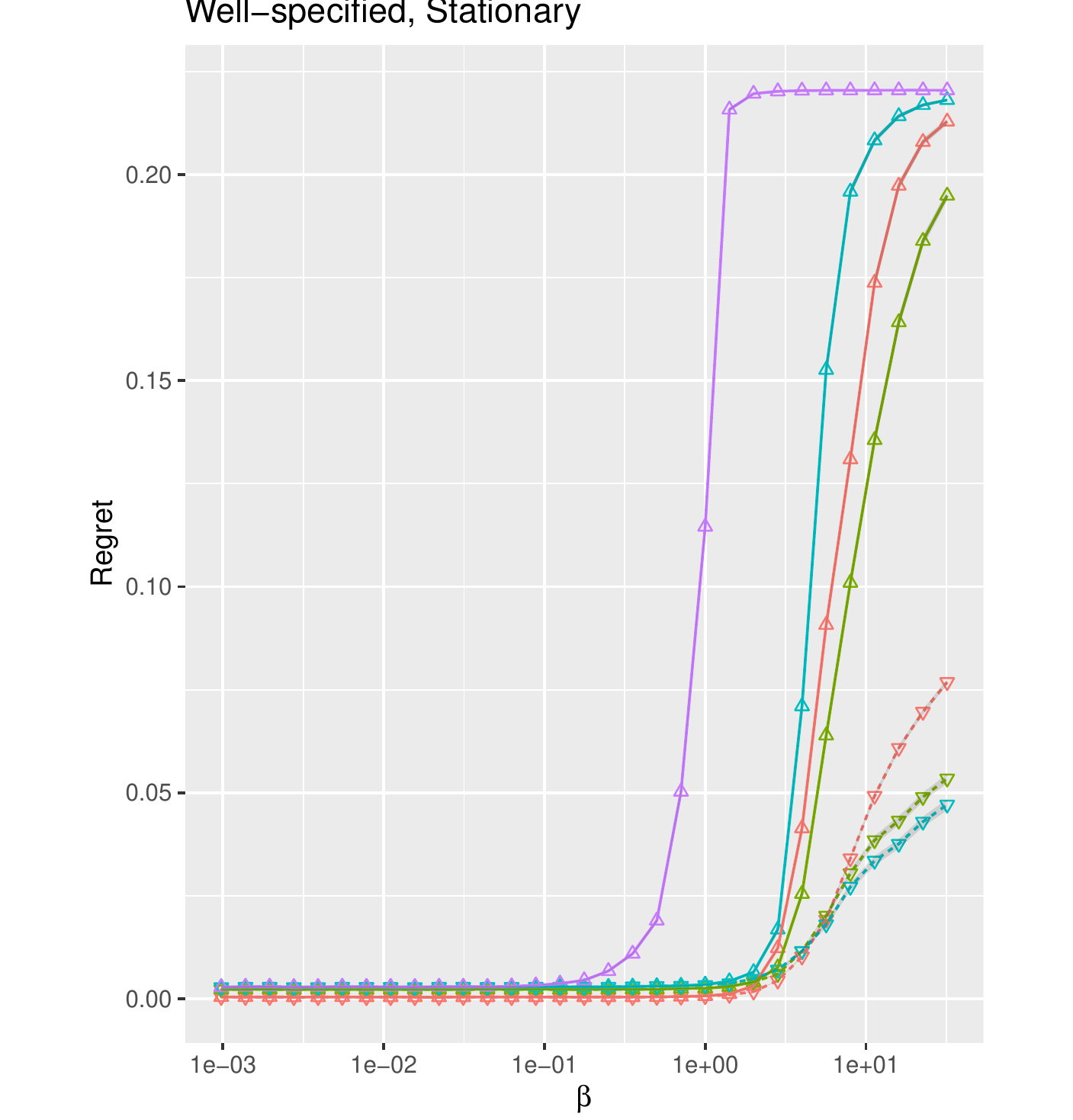}
&
\includegraphics[width=0.29\textwidth]{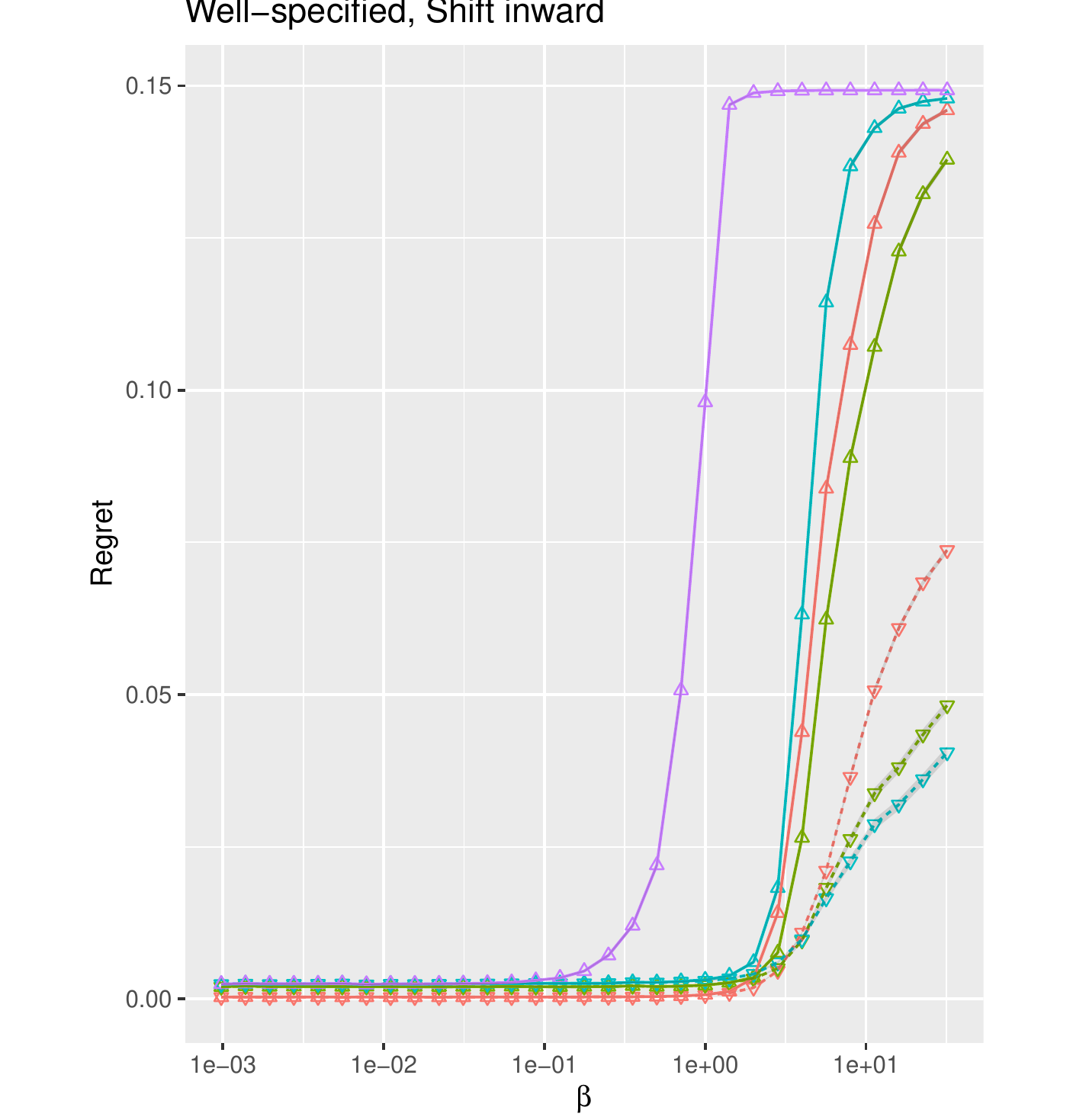}
&
\includegraphics[width=0.29\textwidth]{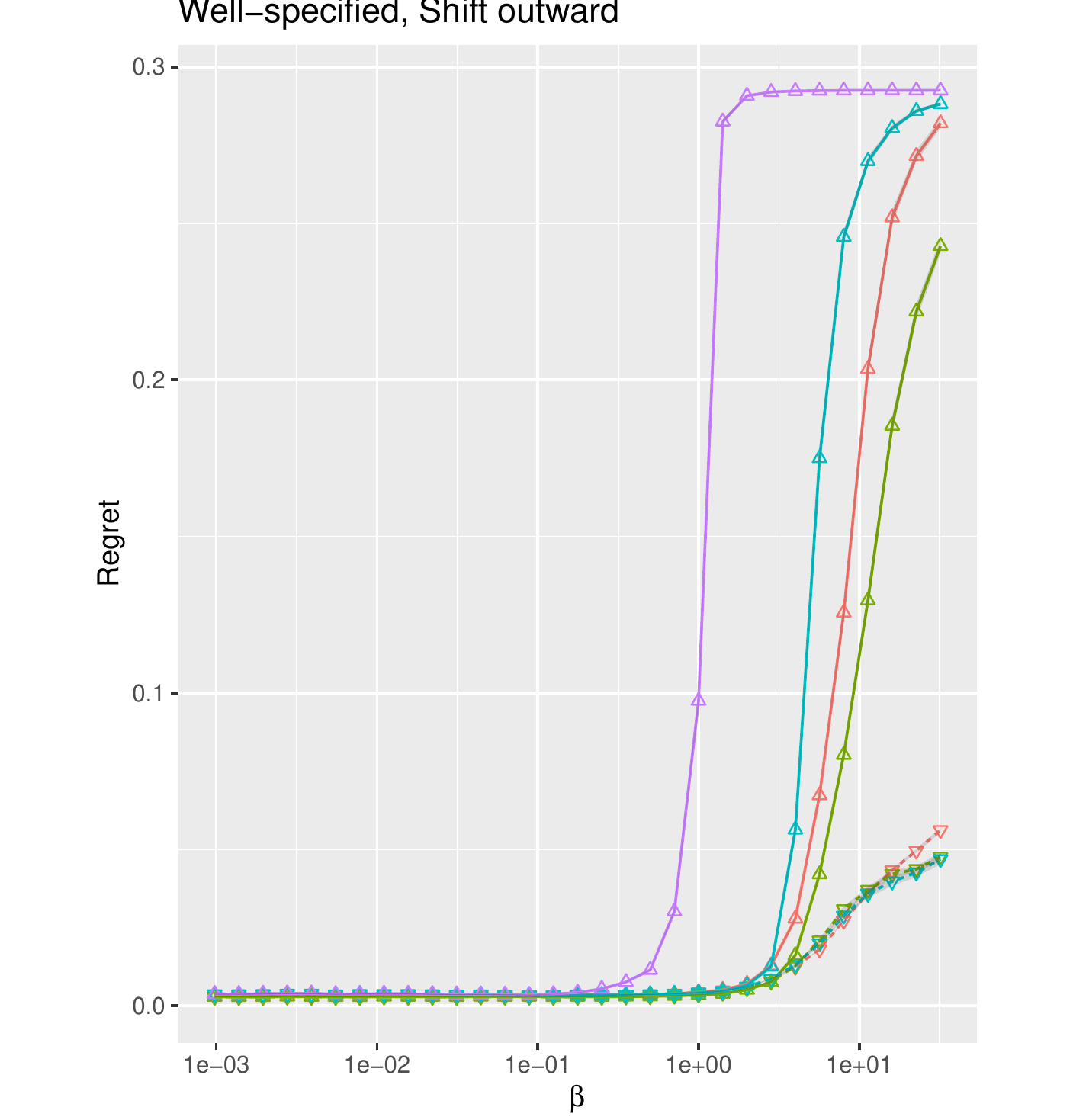}
&
\multirow{-10}{*}[6em]{\hspace{0.025\textwidth}\includegraphics[width=0.125\textwidth]{figures/legend.pdf}}
\\[0.5em]
\includegraphics[width=0.29\textwidth]{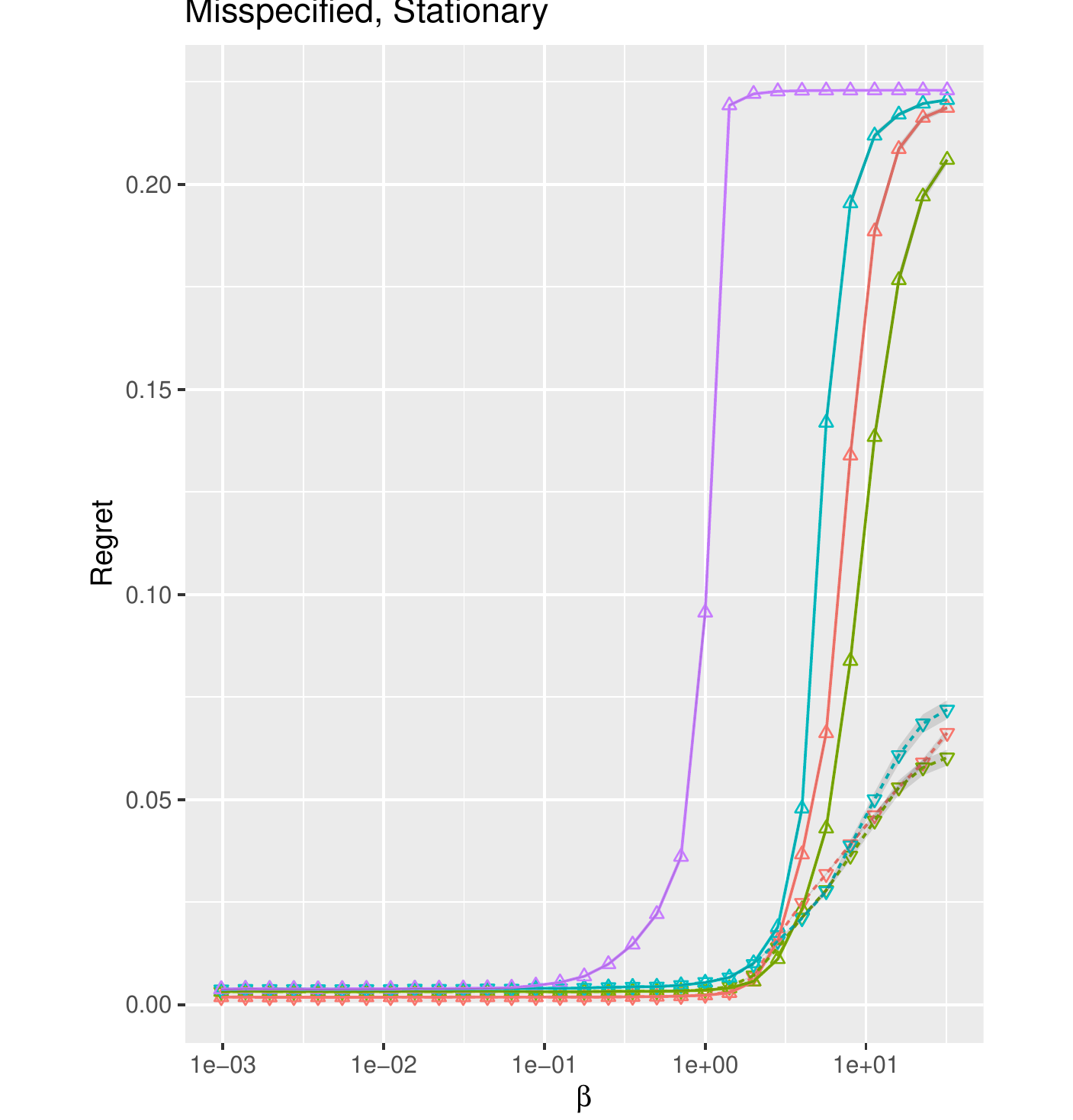}
&
\includegraphics[width=0.29\textwidth]{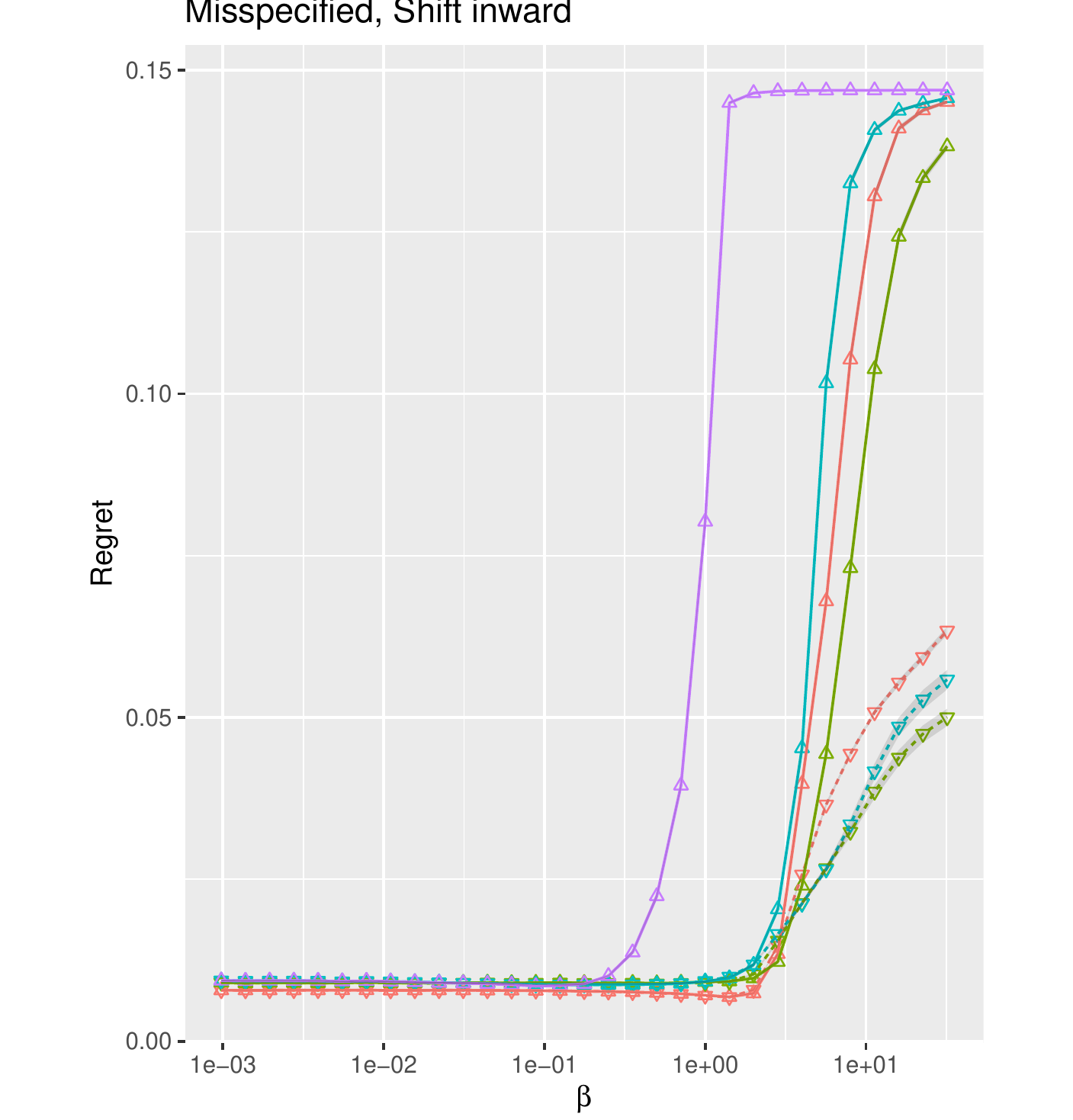}
&
\includegraphics[width=0.29\textwidth]{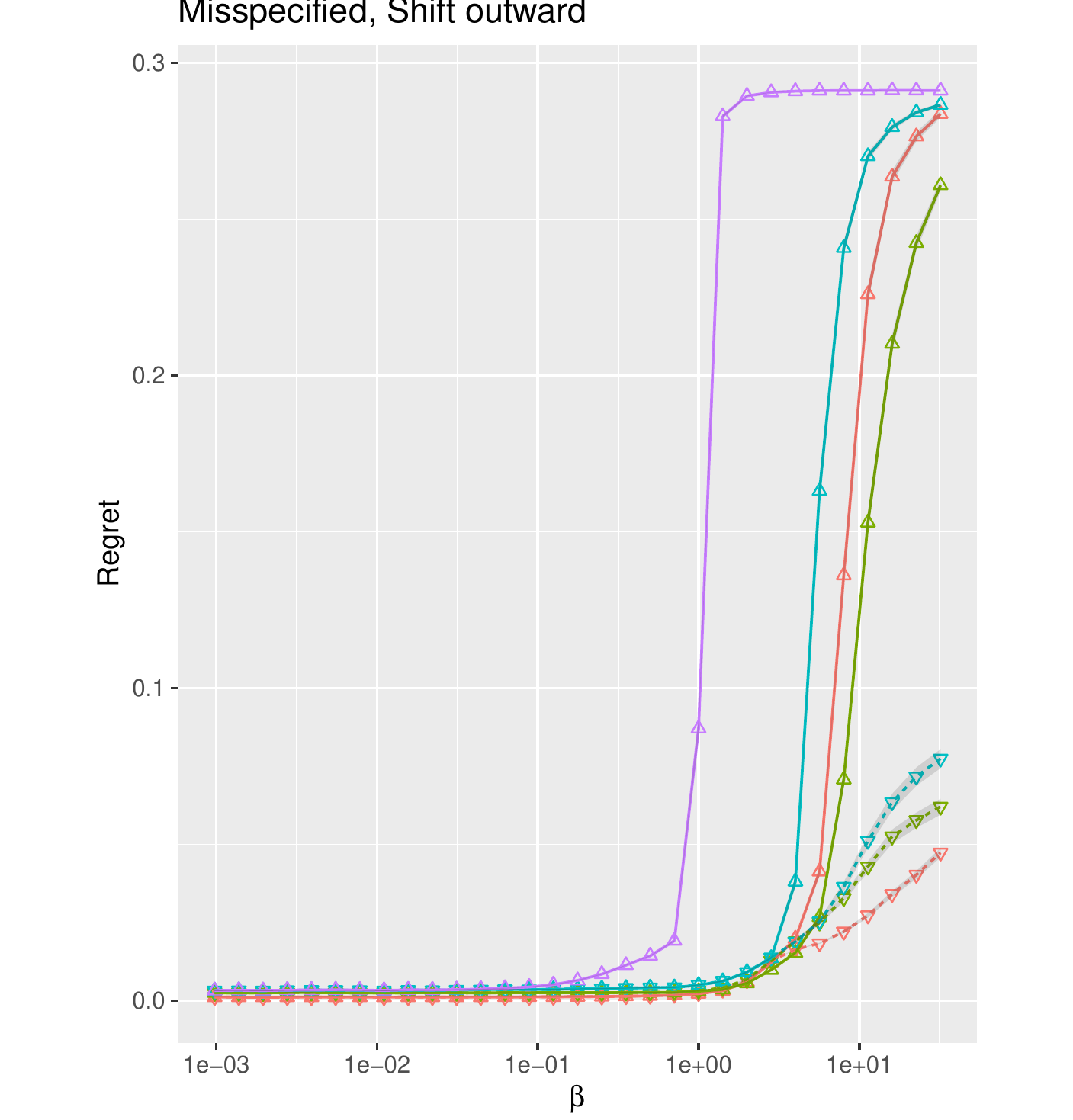}
\end{tabular}%
\caption{Average regret as the level of overlap $\beta$ varies from good to bad ($n=10000$)}\label{fig:synthetic_ovlp}
\end{figure}

We will consider a variety of approaches for learning a deterministic linear policy from $n$ observations from the above data generating process. All of them will take the form of first computing some numbers $\Gamma_1,\dots,\Gamma_n$ and then optimizing
$$
\max_{\theta\in[0,2\pi),\,b\in\Rl}\sum_{i=1}^n\op{sign}(\cos(\theta)X_{i1}+\sin(\theta)X_{i2}-b)\Gamma_i.
$$
In order to avoid any additional layer of complexity that may obscure the conclusions of the numerical results, we solve the above \emph{exactly} by using grid search rather than use a convex surrogate classification loss. Specifically, we conduct a grid search on $\theta$, sort the data along $\cos(\theta)X_{i1}+\sin(\theta)X_{i2}$, and consider $b$ being every midpoint along with $-\infty,\infty$.

Given a training dataset,
we first estimate $\hat\phi,\hat\mu$ using gradient boosting machines, as implemented in the \emph{R} package \texttt{gbm}. 
Given these estimates,
we consider the following policy learning methods:
\begin{enumerate}[leftmargin=*,,labelindent=0in,itemsep=0ex,partopsep=0ex,parsep=0ex]
\item Unweighted: $\Gamma_i=A_iY_i$.
\item IPW: $\Gamma_i=Y_i/(A_i+\hat\phi(X_i))$.
\item Direct: $\Gamma_i=\hat\mu(+\mid X_i)-\hat\mu(-\mid X_i)$.
\item DR: $\Gamma_i=\hat\mu(+\mid X_i)-\hat\mu(-\mid X_i)+(Y_i-\hat\mu(A_i\mid X_i))/(A_i+\hat\phi(X_i))$.
\item Retargeted: for each of the above, we modify $\Gamma_i\gets(1-\hat\phi^2(X_i))\Gamma_i$.
\end{enumerate}
For numerical stability, we normalize all of these by $\eavg\abs{\Gamma_i}$ before solving the above optimization problem. 

\Cref{fig:synthetic_n} shows the average regret as we vary the training dataset size $n$ in the various scenarios \edit{(we run 5000 replicates for each of 20 $n$ values evenly spaced on the logarithmic scale of the plot axis and report averages and standard errors within each $n$ value)}. In the well-specified scenarios we see that the retargeting significantly improves policy learning, achieving an order of magnitude lower regret (note log scale).
In the misspecified scenarios, retargeting still significantly improves performance in moderate samples. In very large samples with misspecification, the improvement is diluted due to hitting an approximation error limit. In particular, at the rightmost end of the plot, we see the regret of retargeted policy learning reaching a floor and slightly overtaken by the non-retargeted DR method, which flattens out as well due to the misspecification. The retargeted versions of other methods still uniformly improve on their non-retargeted versions in this regime.
\Cref{fig:synthetic_ovlp} shows the regret as we vary the level of overlap from perfect, $\beta=0$, to perfectly bad, $\beta\to\infty$
\edit{(we run 5000 replicates for each of 30 $\beta$ values, evenly spaced on the logarithmic scale of the plot axis)}.
We first note that while the unweighted method is indeed consistent when $\beta=0$, it succumbs to confounding bias quickly as $\beta$ grows. The non-retargeted methods account for the confounding but succumb to lack of overlap for slightly larger $\beta$. The retargeted methods account both for confounding and lack of overlap and, while they of course suffer too from the worsening overlap, they consistently perform much better than their non-retargeted versions. (Note that for \cref{fig:synthetic_n} we chose a moderate level of overlap, $\beta=3.5$, where the effect is most nuanced. \Cref{fig:synthetic_ovlp} shows that the effect would be uniformly lopsided in favor of retargeting when overlap is particularly bad, \eg, $\beta=10$.)

The conclusion from these experiments is that retargeting offers an easy way to significantly improve policy learning in the absence of good overlap, and that it is generally robust to both correct specification, which underlied \cref{lemma:weightinvariance} and our motivation for retargeting, and to covariate shifts.

\subsection{Case Study: Multi-Action Personalized Job Counseling}

We next consider an application to a dataset derived from a large scale experiment comparing different programs offered to unemployed individuals in France \citep{behaghel2014private}.
The experiment compared three arms: a control arm where individuals receive the standard services provided by the Public Employment Services, a public treatment arm where individuals receive an intensive counseling program run by a public agency, and a private treatment arm with a similar program run by a private agency. 
The hypothetical application we consider is the learning of a personalized intervention policy to efficiently allocate the resources of the intensive treatments to maximize the number of unemployed individuals who reenter employment within six months, minus costs. 
(The original study focused instead on the differences of the public and private programs, their cost effectiveness, and the incentive problems in contracting the private provider.)

Some individuals assigned to the two treatment arms refused the additional program and reverted to control. Nonetheless, we restrict our attention to intent-to-treat interventions; that is, we consider the actions of our policy as the offer (or no offer) of an intensive program of either type. This make sense since our personalized interventions may not force individuals into a program, only allocate access to the resource. There are of course important fairness considerations in offering differential access to public resources. \citet{kallus2019assessing} study how to assess disparate impact metrics for personalized interventions despite fundamental unidentifiability issues. They also study this same dataset.
Here, for the sake of focus, we only consider the social welfare of policies and ignore fairness or equity considerations.
Furthermore, to make our policies focus on heterogeneous effects, we set the costs of each arm to be equal to their within-arm average outcome in the original data.
That is, the outcome we consider is equal whether one reentered employment within 6 months, minus the average number of individuals who entered employment within 6 months in that arm.
The covariates we consider personalizing on are:
statistical risk of long-term unemployment,
whether individual is seeking full-time employment,
whether individual lives in sensitive suburban area,
whether individual has a college education,
the number of years of experience in the desired job, and
the nature of the desired job (\eg, technician, skilled clerical worker, etc.).

To evaluate the different policy learning methods, we run 1440 replications of the following procedure. Out of the 43977 individuals in the study, we remove a random subset of 8795 (20\%) for training; the remaining individuals will be used for evaluation. We then introduce some confounding into the training dataset. We consider the following three binary variables: whether individual has 1--5 years experience in the desired job, whether they seek a skilled blue collar job, and whether their statistical risk of long-term unemployment is medium. After studentizing each variable, we segment the data by the tertiles of their sum. In the first tertile, we drop each individual in the control arm with probability $1/4$ and in the private and public treatment arms with probability $7/8$. In the second and third tertiles, we repeat this, exchanging the probability for the control arm with that for the private and public treatment arms, respectively. This now serves as our training data. 

We will consider a variety of approaches for learning a linear policy on this data. All of them will take the form of first computing some numbers $\Gamma\in\R{n\times\A}$ and then optimizing
\begin{align*}
\max\quad&\sumn\sumA z_{ia}\Gamma_{ia}\\
\text{s.t.}\quad&b\in\R{\A},\,\beta\in[-1,1]^{\A\times\X},\,z\in\{0,1\}^{n\times\A},\\
&\sumA z_{ia}=1&&\forall i=1,\dots,n\\
&(z_{ia}=1)\implies(b_a+\beta_a^TX_i\geq b_{a'}+\beta_{a'}^TX_i)&&\forall i=1,\dots,n,\,\forall a\neq a'\in\A.
\end{align*}
Again, to avoid the additional complexity of relaxations that may obscure the results, we solve the above as a mixed-integer optimization problem using Gurobi (we put a time limit of 5 minutes on each call).

We next estimate $\hat\phi,\hat\mu$ using gradient boosting machines, as implemented in the \emph{R} package \texttt{gbm}. 
Given these estimates,
we consider the following policy learning methods:
\begin{enumerate}[leftmargin=*,,labelindent=0in,itemsep=0ex,partopsep=0ex,parsep=0ex]
\item IPW: $\Gamma_{ia}=\indic{A_i=a}Y_i/\hat\phi(a\mid X_i)$.
\item Direct: $\Gamma_{ia}=\hat\mu(a\mid X_i)$.
\item DR: $\Gamma_{ia}=\hat\mu(a\mid X_i)+\indic{A_i=a}(Y_i-\hat\mu(a\mid X_i)/\hat\phi(a\mid X_i)$.
\item Retargeted: for each of the above, we modify $\Gamma_{ia}\gets(\sum_{a\in\A}\hat\phi^{-1}(a\mid X_i)+\frac12)^{-1}\Gamma_{ia}$.
\end{enumerate}

Given a learned policy, we evaluate it on the held-out test set using a Horvitz-Thompson estimator (note that the randomized study had varying randomization probabilities; we use these in the evaluation).
We report the results in \cref{table:jobtrain}, showing both average value as well as standard errors \edit{over replications}.
We find that DC and DM perform particular badly suggesting that the outcome model was a bad fit, which may also contribute to DR's worse performance than IPW. Nonetheless, for both DR and IPW, retargeting offers a clear and substantial improvement in policy value.

\begin{table}[t!]\centering
\setlength{\tabcolsep}{1em}%
\begin{tabular}{lccc}\toprule
       & \multicolumn{2}{c}{Average Policy Value (in 1000's)} \\
Method & Conventional & Retargeted & Improvement \\\midrule
DC & $-3.44\pm0.005$ & --- & ---\\
DM & $-3.42\pm0.005$ & $-3.42\pm0.005$ & 0\%\\
DR & $\phantom{-}1.14\pm0.006$ & $\phantom{-}1.93\pm0.006$ & 70\% \\
IPW & $\phantom{-}2.09\pm0.006$ & $\phantom{-}2.65\pm0.005$ & 27\%\\\bottomrule
\end{tabular}
\caption{Average policy values of different policy learning methods applied to the job counseling dataset, with standard errors \edit{over replications}.}\label{table:jobtrain}
\end{table}

\section{Discussion}\label{sec:discussion}

{\blockedit\subsection{The Choice of Efficiency Bound to Optimize}\label{sec:discussion efficiency bound}

In our method, we focused on minimizing the efficient asymptotic error of estimating the retargeted sample regret, ${R_n}(\pi;w,\rho)$.
Notably, this asymptotic error was only {part} of the efficient asymptotic error of estimating the retargeted population regret, ${R}(\pi;w,\rho)$, which also included the term in \cref{eq:decomposeefficientvariance1}. In this section we offer further discussion of this choice.

In certain senses, ${R_n}(\pi;w,\rho)$ is the best finite-sample objective we can hope for. It completely eliminates any issues of residual noise and, more importantly, any issue of lack of overlap in the data. 
Beyond residual noise and overlap issues, any additional error may arise only from the sampling uncertainty of the finite dataset of $X$ values. Recall that ${R_n}(\pi;w,\rho)$ is measurable with respect to just $X_{1:n}$. Indeed, the remaining term in the efficiency bound for ${R}(\pi;w,\rho)$, \cref{eq:decomposeefficientvariance1}, is the $X$-variance of $R_n(\pi;w,\rho)$. This term of the efficiency bound is driven solely by heterogeneity in $X$.

Our primary justification for omitting this term from consideration is that, in view of \cref{lemma:weightinvariance2}, under correct specification, the choice of $w$ does not affect the optimizer of $R_n(\pi;w,\rho)$ and therefore it does not matter which $w$ makes $R_n(\pi;w,\rho)$ best estimate the population quantity $R(\pi;w,\rho)$, the result is always the same. What matters is estimating $R_n(\pi;w,\rho)$. We therefore focused on \cref{eq:decomposeefficientvariance2}, the efficiency bound for this latter estimation.

Under misspecification this argument may not hold, but this is true to the same extent that retargeting may be inappropriate to begin with. We discuss the issue of misspecification further in \cref{sec:discussion misspecification and shift} below.
Note that since our data arrives from the uniformly-weighted population, using a non-constant $w$ would usually (but not necessarily) inflates \cref{eq:decomposeefficientvariance1} (in particular, the minimum such term is given by the non-constant $w\propto\prns{\sumA(\pi(a\mid X)-\rho(a\mid X))\mu(a\mid X)}^{-1}$).
At any rate, minimizing the $X$-heterogeneity term would have us focus on populations with very little heterogeneity, while heterogeneity is exactly what we focus on and exploit in policy learning. 

Finally, we point out that if we know on which $X$-population we wish to evaluate regret, then the efficiency bound for the population quantity is exactly \cref{eq:decomposeefficientvariance2}, which we focused on, without an $X$-heterogeneity term, as the following shows.
\begin{lemma}\label{lemma:efficientinfluence2}
Suppose the measure $\nu$ of $X$ is known.
Then the efficient influence function for $R(\pi;w,\rho)$ is $\psi'(x,a,y;\phi,\mu)-R(\pi;w,\rho)$ where
\begin{align*}
\psi'(x,a,y;\phi_0,\mu_0)=&\int w(x')\sumA*{a'}(\pi(a'\mid x')-\rho(a'\mid x'))\mu_0(a'\mid x')\;d\nu(x')\\&+w(x)\frac{\pi(a\mid x)-\rho(a\mid x)}{\phi_0(a\mid x)}(y-\mu_0(a\mid x)).
\end{align*}
And, the efficiency bound for $R(\pi;w,\rho)$ in this case is
given by \cref{eq:decomposeefficientvariance2}.
\end{lemma}
If we imagine instantiating the above with $\nu$ being the empirical distribution, then, in a sense, this is another way of stating that the ``efficiency bound'' for estimating ${R_n}(\pi;w,\rho)$ is given by \cref{eq:decomposeefficientvariance2}. (Note that if $\nu$ is the empirical distribution then the sample average of $\psi(X,A,Y;\phi_0,\mu_0)$ is the same as that of $\psi'(X,A,Y;\phi_0,\mu_0)$.)
More generally, this shows that if we know the population distribution of $X$, then the efficiency bound for estimating the \emph{population} regret, $R(\pi;w,\rho)$, is simply \cref{eq:decomposeefficientvariance2}. (Note that in this case the structure of an efficient estimator changes to depend on $\nu$ as in $\psi'(X,A,Y;\phi_0,\mu_0)$ above.)%
}

\subsection{Retargeting, Misspecification, and Train-Test Covariate Shift}\label{sec:discussion misspecification and shift}
In \cref{sec:invariance}, we argued that if $\Pi_0$ is well-specified then we may arbitrarily change the distribution of covariates, and retargeted policy learning will still give the unconstrained best policy. If $\Pi_0$ is misspecified, however, retargeted policy learning will instead give the policy in $\Pi_0$ that is best on average for the particular retargeted population, which may be different from the best policy in $\Pi_0$ that is best for the original training population. {\blockedit In particular, if we use the optimal retargeting weights then we will get the policy in $\Pi_0$ that is best on the regions where no action has a small propensity $\phi(a\mid X)$, what \citet{li2018balancing} call ``marginal'' units in the binary case. Under correct specification, this is the same as the optimal policy, while under misspecification, this policy depends on the treatment mechanism.}

However, in practice, one can inevitably expect some amount of covariate shift between the training data and the test data where the policy will be deployed. Usually, we have training data from a particular study or from the subset of records with sufficient information, we learn a policy on this data, and we intend to deploy it on a larger scale. And usually the exact test distribution is not known a priori (if it is, we can target it; see, \eg, \citealp{zhao2019robustifying} \edit{and \cref{lemma:efficientinfluence2}}). In light of this, it may seem overly dogmatic to insist on getting the best $\Pi_0$ policy for the training distribution of covariates \editb{for fear of misspecification}, especially when it can have very bad overlap \editb{and misspecification potentially mild}.
Therefore, retargeting may still be preferable even under misspecification. Indeed, we saw in \cref{sec:sim} that even under misspecification, with or without any covariate shift, retargeting offered better performance in all reasonably sized datasets.
\edit{We can also attenuate misspecification effects by making our policy class more flexible or by regularizing retargeting weights as in \cref{sec:biasreg}.}
In the end one must consider and compare the level of misspecification, the level of inevitable (and unknown) train-test covariate shift, and the level of overlap. Given all these interacting uncertain factors, we think it is both safest and most effective to just focus on learning a policy on a subpopulation with good overlap, where the policy learning procedure is actually reliable and robust. That is, we think retargeting is generally a safe bet and should be applied generally.

\subsection{On the Interplay of Retargeting and Regret}\label{sec:dicussion regret}
Our approach to retargeting was motivated by the observation that there are several characterizations of the optimal policy as the maximizer of different objectives. We therefore sought the most convenient characterization, with the most efficiently estimable objective. It remains an open question, however, how to characterize the efficiency of learning the optimal policy itself as the parameter of interest, for example when assuming a parametrized form of the policy class such as linear. 

{\blockedit
Following a uniform convergence argument, we could characterize the regret on the same retargeted distribution used for training by retargeted policy learning.
\begin{lemma}\label{lemma:unifconv2}
Under the same conditions as \cref{lemma:unifconv}, there exists a universal constant $C$ such that, for any $\delta\in(0,1/2)$ and $\rho_0\in\R{\A\times X}$, with probability at least $1-\delta$,
\begin{align*}
&\sup_{\pi,\pi'\in\Pi_0}\abs{(V_n(\pi;w)-V_n(\pi';w))-(V(\pi;w)-V(\pi';w))}
\\&\qquad\qquad\leq
C\prns{\kappa(\Pi_0)+1+\sqrt{\log(1/\delta)}}
\sqrt{\frac{\Eb{{w^2(X)M^2(X)}}}n}+o\prns{\frac{\log(1/\delta)}{\sqrt{n}}},
\end{align*}
where $M(x)=\max_{a\in\A}\mu(a\mid x)-\min_{a\in\A}\mu(a\mid x)$.
\end{lemma}
If we have a uniformly efficient estimator, $\sup_{\pi\in\Pi_0}\abs{\hat V_n(\pi;w)-\tilde V_n(\pi;w)}=o_p(1/\sqrt{n})$, then
by combining \cref{lemma:unifconv,lemma:unifconv2} and instantiating them at a maximizer of $V(\pi;w)$ and at any $\hat\pi^w_n\in\argmax_{\pi\in\Pi_0}\hat V_n(\pi;w)$, we obtain \begin{equation}\label{eq:regretbd}\max_{\pi\in\Pi_0}V(\pi;w)-V(\hat\pi^w_n;w)=O_p\prns{(\kappa(\Pi_0)+1)
\sqrt{{\Omega(w,\rho_0)+\Eb{{w^2(X)M^2(X)}}}}/\sqrt{n}}.\end{equation}
This shows that we may obtain better regret on the retargeted population.
However, this is limited to that population, as could be expected as the result holds regardless of correct specification.
While under correct specification we have $\max_{\pi\in\Pi_0}V(\pi;w')-V(\pi';w')\leq (\esssup w'(x)/w(x)) (\max_{\pi\in\Pi_0}V(\pi;w)-V(\pi';w))$ for any $w,w'\in\R{\X}_++$, this is still too loose to characterize the regret on the original population ($w'=\bm1$) and washes out the regret improvements we actually see in practice due to retargeted policy learning. (While in view of the previous section, if there were a covariate shift, then this might be helpful as the ratio between the test distribution and training need not be smaller than that between the test and the retargeted population.) The bound also depends on the second moment of $V_n(\pi;w)$, while the set of maximizers of the latter is the same for all $w$, which was our motivation for ignoring this term in our efficiency bound objective. (Although, if the distribution of $X$ were known, then 
via \cref{lemma:efficientinfluence2} we could also obtain \cref{eq:regretbd} \emph{without} the second variance term.)

Instead, perhaps a better lens through which to understand the regret improvements due to retargeted policy learning is that we learn the underlying optimal policy parameters faster, and can therefore generalize better to any population, including the original one. As stated above, it remains an open question outside our scope how to characterize this in generality. Below we give a primitive treatment of this for the case of a finite class of policies to illustrate the idea.
\begin{lemma}\label{lemma:finitepolicyconvergence}
Let $p_\pi=\Prb{\pi(a\mid X)(\max_{a'\in\A}\mu(a'\mid X)-\mu(a\mid X))=0\;\forall a\in\A}$ and $p=\sup_{\pi\in\Pi_0:p_\pi\neq 1}p_\pi$. Suppose $\Pi_0\cap\Pi^*\neq\varnothing$, $w\in\R{\X}_{++}$.
Define $\Pi_{0,n}^*(w)=\argmax_{\pi\in\Pi_0}V_n(\pi;w)$. Then,
$$
\Prb{\Pi_{0,n}^*\not\subseteq\Pi^*}=\Prb{\Pi_{0,n}^*(w)\not\subseteq\Pi^*}
\leq \abs{\Pi_0}p^n.
$$
\end{lemma}
Now consider any $\hat\Pi_n$ and notice that
$$
\fPrb{\hat\Pi_n\not\subseteq\Pi^*}
\leq\fPrb{\hat\Pi_n\not\subseteq\Pi_{0,n}^*(w)}+\fPrb{\Pi_{0,n}^*(w)\not\subseteq\Pi^*}.
$$
The latter probability is exponentially decaying in $n$ if $\abs{\Pi_0}<\infty$ and in a manner independent of $w$, per \cref{lemma:finitepolicyconvergence}. 
The former probability has to do with how well we are approximating the problem of optimizing $V_n(\pi;w)$, as characterized by, \eg, \cref{lemma:unifconv}. In particular, focusing on $\hat\Pi_n=\argmax_{\pi\in\Pi_0}\tilde V_n(\pi;w)$, we have:
\begin{lemma}\label{lemma:finitepolicyconvergence2}
Suppose that $\phi(A\mid X)$ is bounded away from zero and $Y$ is bounded.
Then
there exist $\alpha>0,\,\beta<1$ such that for any $n\geq1,w\in\R{\X}_{++},\rho\in\R{\A\times \X}$, 
$$
\Prb{
\argmax_{\pi\in\Pi_0}\tilde V_n(\pi;w)
\not\subseteq\Pi_{0,n}^*(w)}
\leq
2\abs{\Pi_0}\E\exp\prns{\frac{-n\min(\gamma^2_n(w),\alpha)}{32\Omega(w,\rho)}}
+\abs{\Pi_0}\beta^n,
$$
where
$\gamma_n(w)=\max_{\pi\in\Pi_0} V_n(\pi;w)-\max_{\pi\in\Pi_0\backslash\Pi_{0,n}^*(w)} V_n(\pi;w)$.
\end{lemma}
Thus, we see that the probability we successfully pick out an optimal policy in $\Pi^*$ improves when we make $\Omega(w,\rho)$ small. However, it also depends on the separation that a particular $w$ induces on the empirical value of different policies, $V_n(\pi;w)$. Indeed some $w$'s may make it small and some large. However, a priori there is no reason to believe that this separation is smaller for the original population ($w=\bm1$) than for any other retargeted population, such as the optimal one ($w=w_0$), and therefore we expect to do better when $\Omega(w,\rho)$ is small. (This is unlike the variance of $V_n(\pi;w)$, which we do expect to be inflated; see \cref{sec:discussion efficiency bound}.) This illustrates the regret benefits of optimizing a retargeted objective with small $\Omega(w,\rho)$, but a general characterization remains open.%

\editb{Much of the complexity here arises from the non-regularity of policy learning \citep{laber2014dynamic}. The benefit of retargeting is more straightforward for regular estimands. Consider estimating the linear projection of some $Z$ onto $X$, \ie, $\beta^*$ solving $\E[(Z-\beta^TX)X]=0$. Without additional information, ordinary least squares is the best one can do.
Consider, for example, doing instead weighted least squares weighted by $1/\op{Var}(Z\mid X)$. Generally, this identifies a \emph{different} projection from $\beta^*$.
If, however, $\E[Z\mid X]$ were linear, then it identifies the same $\beta^*$ and it increases estimation efficiency, which in turn also translates to lower squared-error risk due its bowl-shaped loss \citep[Lemma 8.5]{van2000asymptotic}.
This is immediately applicable to personalized causal inference. 
If $\A=\{-,+\}$ and $Z=\psi(X,A,Y;\phi,\mu)$ ($\psi$ as in \cref{lemma:efficientinfluence}), then $\E[(Z-\beta^TX)X]=0$ is the efficient version of the (unobservable) estimating equation $\E[(Y(+)-Y(-)-\beta^TX)X]=0$ \citep{tsiatis2007semiparametric}. {If} $\E[Y(+)-Y(-)\mid X]$, known as the conditional average treatment effect (CATE), were linear, then we would be better off reweighting using $1/\op{Var}(Z\mid X)\propto\prns{\frac{\sigma^2(+\mid x)}{1+\phi(x)}+\frac{\sigma^2(-\mid x)}{1-\phi(x)}}^{-1}$, which matches $w_0$ in \cref{lemma:twoarmomega}. Recently, \citet{bennett2020efficient} considered efficient estimation of the policy parameters that solve smooth convex relaxations of the policy value, which \emph{are} regular.}

\editb{As an aside, note that estimating CATE for the purposes of personalization suffers from \emph{worse} missepcification concerns: not only does a misspecified CATE not transport to new populations, it does not even transport to the training population as it optimizes the wrong loss. (Nonetheless, if we seek the best \emph{unrestricted} policy in all $\Pi$, learning a nonparametric CATE may be inevitable; cf. \citealp{stoye2009minimax})}
}

\subsection{Local Uniform Efficiency Objectives}\label{sec:dicussion local}
For retargeting, we sought the weights and reference policy that minimized the uniform control of efficiency bounds over all $\pi\in\Pi$, \ie, $\Omega(w,\rho)$. But in the end, we only need to estimate the objective for $\pi\in\Pi_0$. Therefore, one possible extension of our approach is to instead minimize $\Omega_0(w,\rho)=\sup_{\pi\in\Pi_0}
\Eb{
w^2(X)\sumA\frac{\sigma^2(a\mid X)}{\phi(a\mid X)}\prns{\pi(a\mid X)-\rho(a\mid X)}^2
}$. \edit{In particular, \cref{lemma:unifconv} continues to hold with $\Omega(w,\rho)$ replaced by this $\Omega_0(w,\rho)$.}

In the case of two actions, ${\A}=\{-,+\}$, one can see that, as long as $\Pi_0$ contains at least one deterministic policy, then $\Omega(w,\rho_0)=\Omega_0(w,\rho_0)$ are the same, where $\rho_0(\pm\mid x)=1/2$. 
So, if one is content to use $\rho_0(\pm\mid x)=1/2$ as the reference policy then the optimal retargeting weights remain the same.
However, unlike the case of $\Omega(w,\rho)$, the optimal $\rho$ in minimizing $\Omega_0(w,\rho)$ may not actually be $\rho_0$ and may generally depend on $\Pi_0$. 
Finding the $\Pi_0$-specific optimal reference policy becomes a difficult non-convex optimization problem. This extension, therefore, while potentially appealing, may be very difficult to both implement and analyze.

\section{Conclusion}
\label{sec:conc}

We studied one countermeasure to make policy learning more robust to limited overlap by simply changing the population on which we optimize the policy, which we termed retargeting. We developed our approach in terms of choosing from among the many equivalent characterizations of the optimal policy the one that is most convenient from the perspective of asymptotic estimation efficiency.
In policy learning, one is simply interested in an effective, well-performing policy; not some unbiased estimate with a clean population interpretation, as in the case of treatment effect estimation. 
This makes policy learning particularly apt for retargeting.
When the policy class is well-specified retargeting incurs no bias, but even if the policy class is misspecified, retargeting can lead to more robust policy learning and, in the end, better performance.
We therefore believe that in practice, retargeting should be used in any policy learning procedure applied to observational data, where overlap will inevitably pose a serious issue. Since retargeting is quite simple to apply, this offers an easy fix for any policy learning algorithm.

{\blockeditb\section*{Acknowledgments}

The author thanks the anonymous reviewers and associate editor for the constructive inputs.}
\if0\blind This material is based upon work supported by the National Science Foundation under Grant No. 1846210.\fi

\bibliographystyle{agsm}
\bibliography{retargetting_rev2}

\newpage

\bigskip
\begin{center}
{\Large\bf Online appendix}\\
\vspace{4pt}{\LARGE\bf More Efficient Policy Learning via\\Optimal Retargeting}\if0\blind\\
\vspace{4pt}{\large Nathan Kallus}\fi
\end{center}

\appendix

{\blockedit
\section{Bias Regularization}\label{sec:biasreg}

In this supplemental section we consider interpolating between optimal retargeting weights and uniform weights to protect against cases of misspecification.
In \cref{sec:optretarget}, motivated by the fact that the asymptotic limit of policy learning is invariant under correct specification, we sought to use retargeting weights that lead to the most easily estimable objective and, in particular, attenuate the issue of overlap. However, when the policy class is \emph{misspecified}, we may be concerned that we may bias our policy learning too much. 

In this section we consider seeking retargeting weights that balance bias and asymptotic efficient variance. Specifically, since only retargeting interacts with misspecification, and centering by a reference policy does not, we focus on the bias in the reweighted policy value, \ie,
$$
\abs{V(\pi)-V(\pi;w)}=\abs{\Eb{(w(X)-1)\sum_{a\in\A}\pi(a\mid X)\mu(a\mid X)}}.
$$
This bias depends on the policy, $\pi$, and the outcome regression function, $\mu$. As we did with the asymptotic efficient variance objective, we will consider a worse-case version of the bias. 
Suppose $\mu(a\mid X)\in L_2$ is square-integrable for each $a\in\A$. Then we have that $\mu_{\max}(X)\in L_2$ is also square-integrable where $\mu_{\max}(X)=\max_{a\in\A}\abs{\mu(a\mid X)}$. Therefore, we consider
$$
B(w;\lambda)=\sup_{\pi\in\Pi,\ \fmagd{\mu_{\max}}_{L_2}\leq\lambda}\abs{V(\pi)-V(\pi;w)}.
$$
Note that $B(w;\lambda)=\lambda B(w;1)$ is clearly homogeneous in $\lambda$. So we focus on just $B(w)=B(w;1)$.
\begin{lemma}\label{lemma:biasworstcase}
$$
B(w)=\sqrt{\Eb{(w(X)-1)^2}}
$$
\end{lemma}
Note that the form in \cref{lemma:biasworstcase} would still follow if we defined bias in the centered values, $R(\pi;w,\rho)$, instead and changed the definition of $\mu_{\max}$ appropriately. The key insight is that, appropriately defined, the worst-case bias will generally take the form of the $L_2$ distance between $w$ and uniform weights.

To balance bias and asymptotic efficient variance, we consider choosing weights to minimize a new combined criterion:
$$
\mathcal E^2_\lambda(w)=\lambda^2 B^2(w)+\inf_\rho\Omega(w,\rho).
$$
\begin{lemma}\label{lemma:biasregularization}
Let $\kappa$ be as in \cref{lemma:multiarmomega}. Let $$w_0(x)\propto(\kappa(x)+4\lambda^2)^{-1}$$ such that $\Eb{w_0(X)}=1$. Then $w_0\in\argmin_{w\;:\;\Eb{w(X)}=1}\mathcal E^2_\lambda(w)$.
\end{lemma}
In other words, \cref{lemma:biasregularization} shows that the optimal weights that balance bias and asymptotic efficient variance are given by simply \emph{padding} the weights given the previous sections.
Essentially, this shrinks the optimal retargeting weights toward uniform weights.

In the case of two actions, $\A=\{-,+\}$, and homoskedastic noise, $\sigma^2(\pm\mid x)=\sigma^2$, this can be simplified to
$$
w_0(x)\propto \frac{1-\phi^2(X)}{c+1-\phi^2(X)},
$$
where $c=\frac{\sigma^2}{2\lambda^2}$ is a transformed weight padding parameter.

\subsection{Empirical Results}

\begin{figure}[t!]\centering
\def\arraystretch{0}%
\setlength{\tabcolsep}{-.25em}%
\begin{tabular}{ccc}
\includegraphics[width=0.29\textwidth]{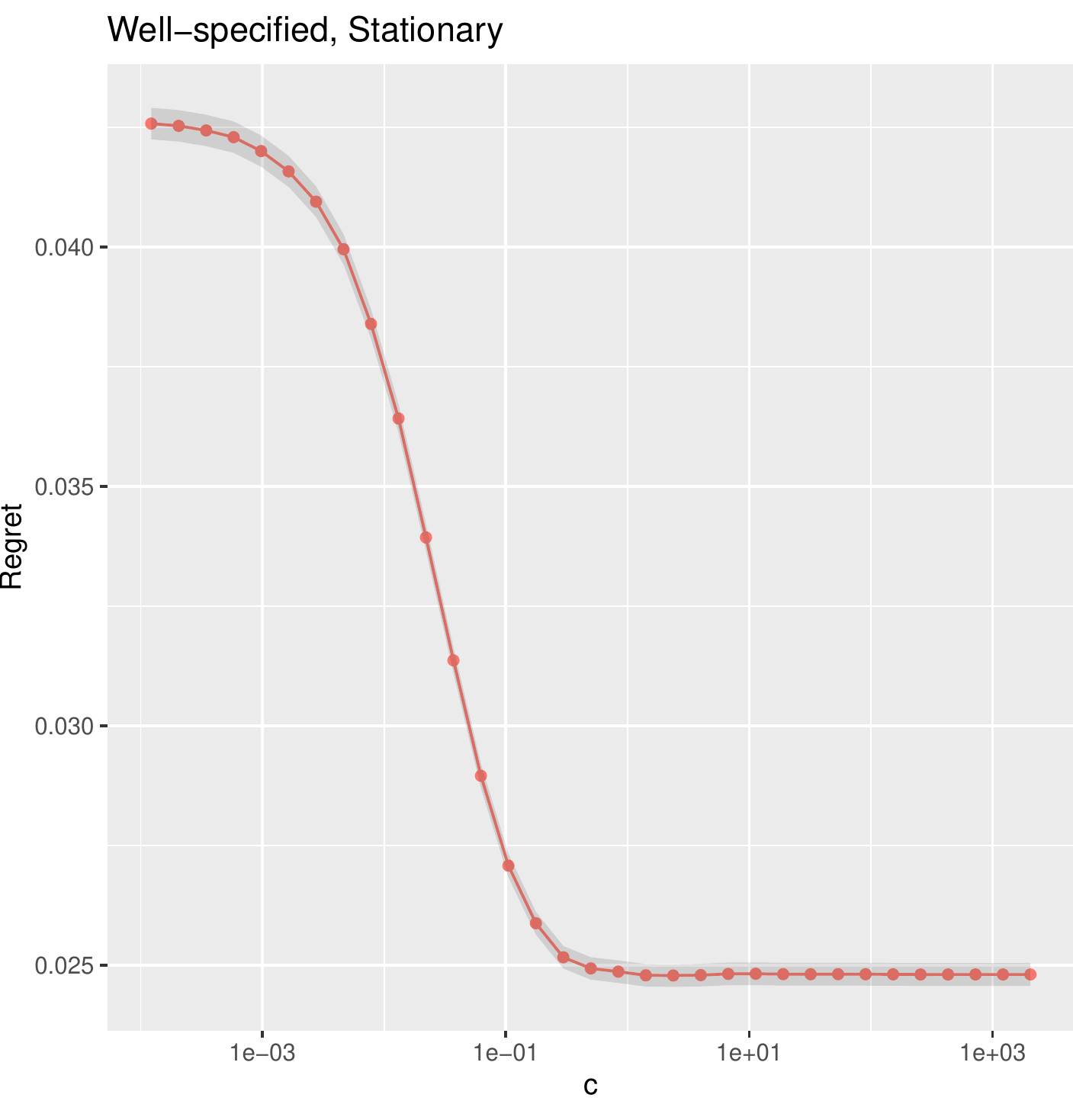}
&
\includegraphics[width=0.29\textwidth]{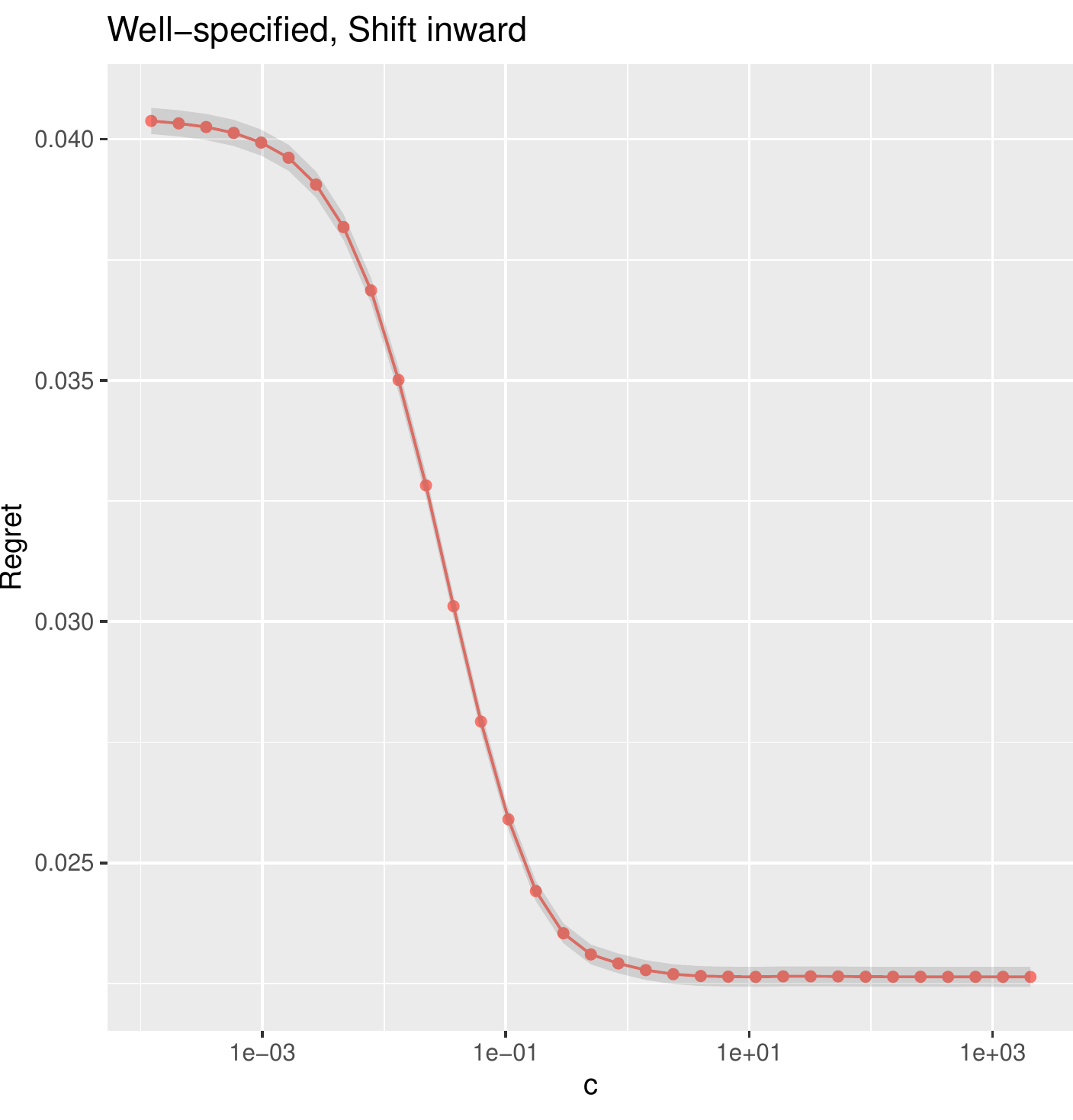}
&
\includegraphics[width=0.29\textwidth]{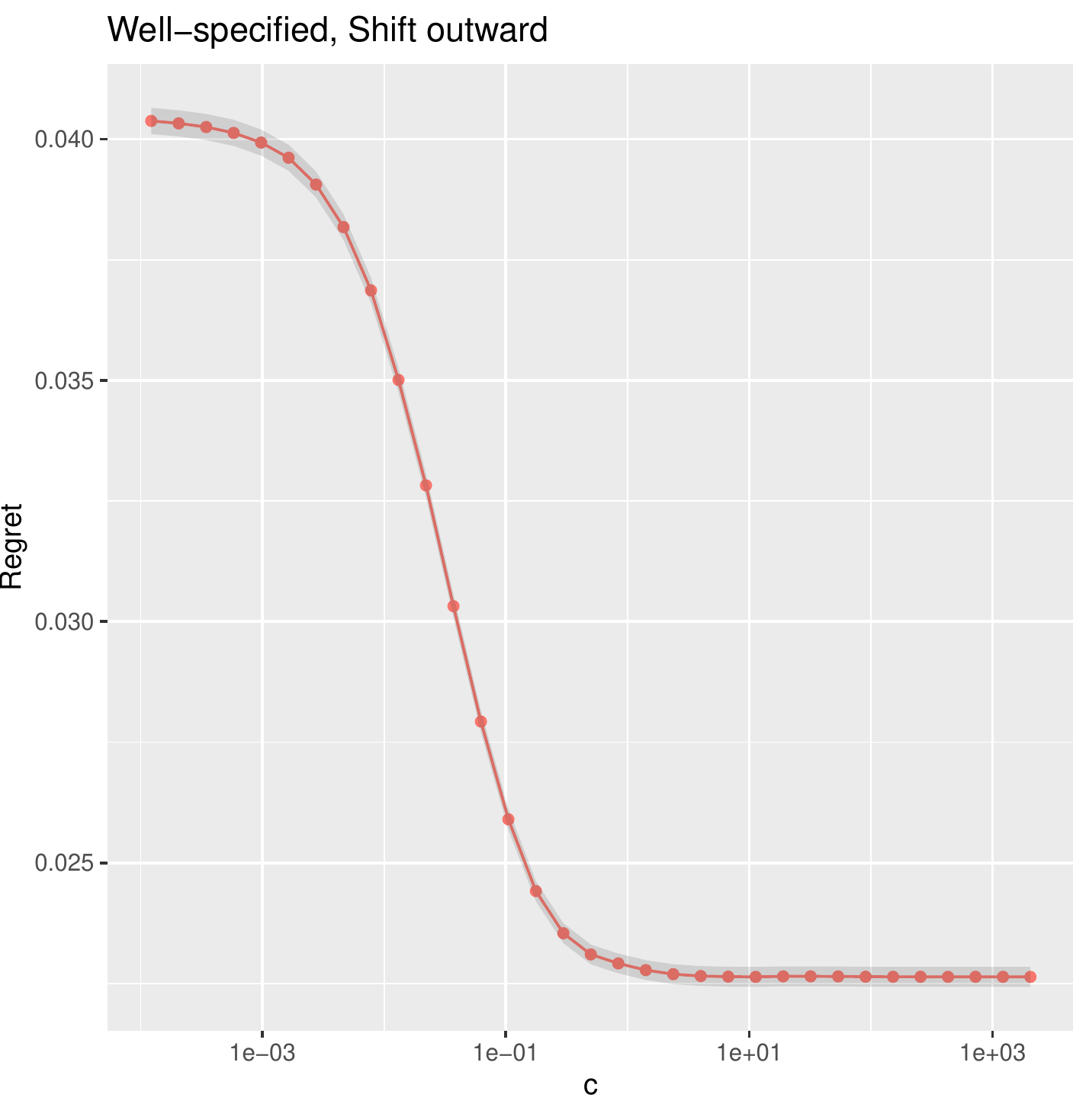}
\\[0.5em]
\includegraphics[width=0.29\textwidth]{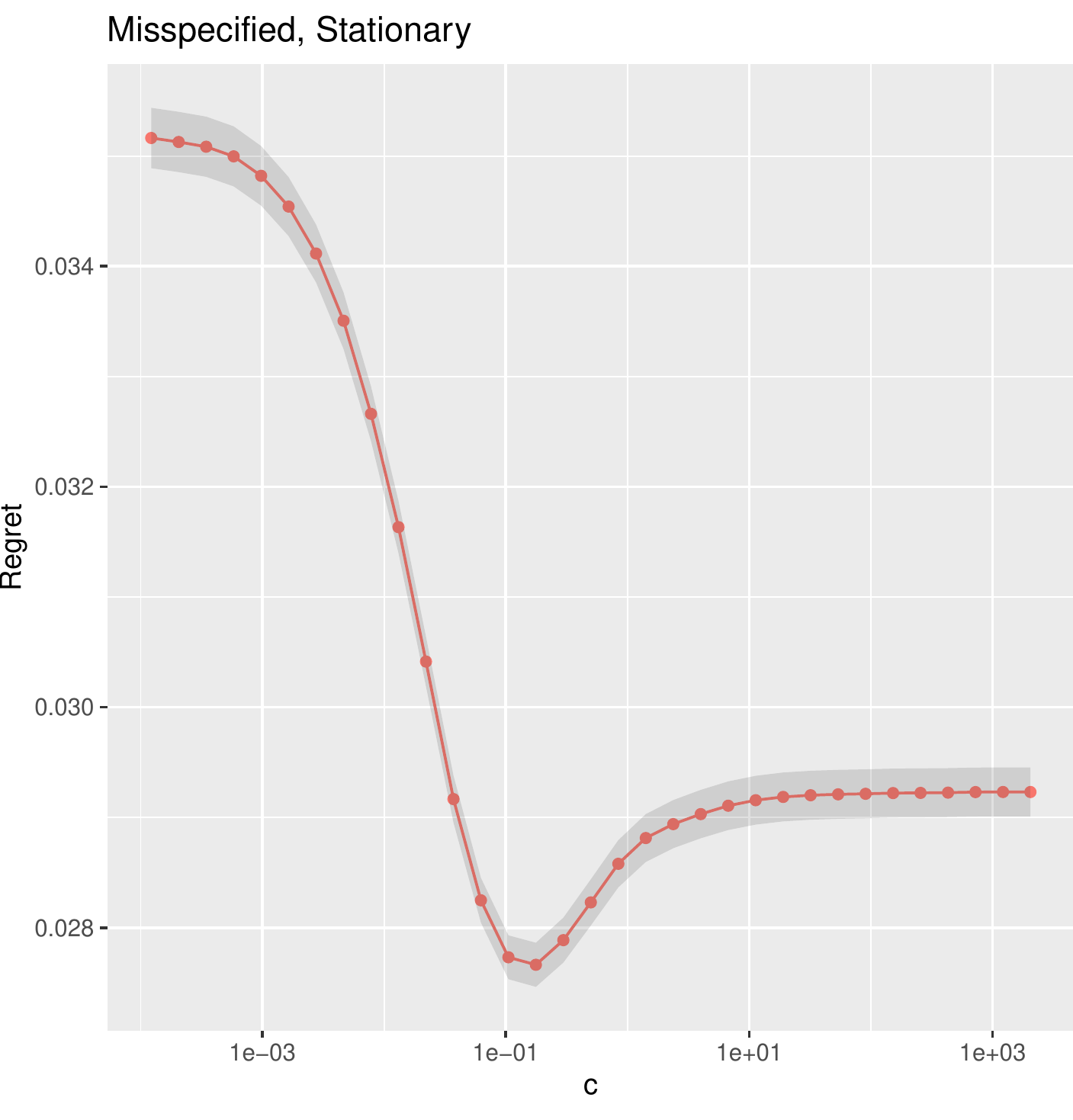}
&
\includegraphics[width=0.29\textwidth]{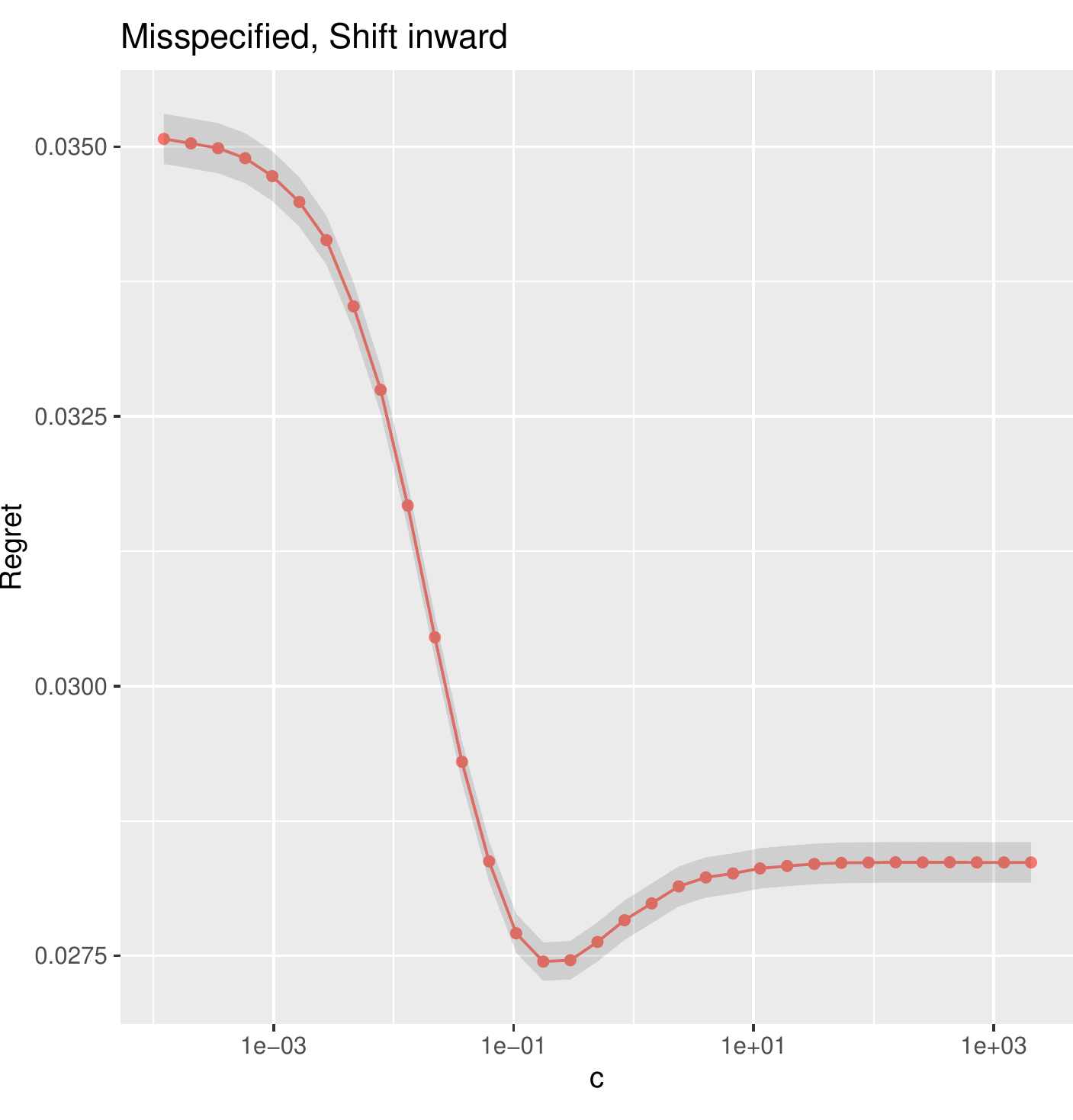}
&
\includegraphics[width=0.29\textwidth]{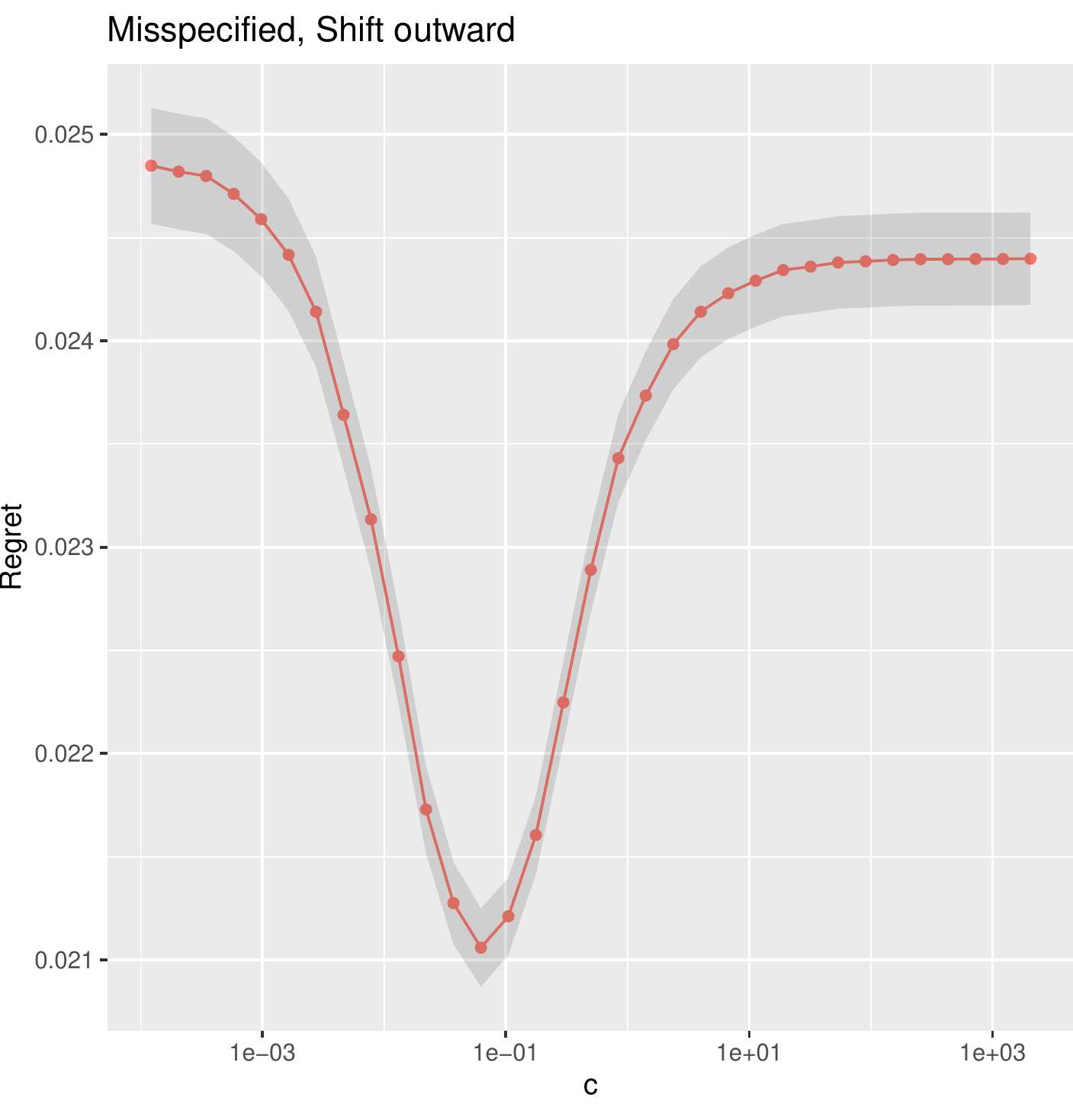}
\end{tabular}%
\caption{\edit{Average regret as we vary retargeting level $c$ from none to full ($n=2000$)}}\label{fig:synthetic_pad}
\end{figure}

To study the impact of bias regularization, we repeat the experiential setup of \cref{sec:sim} and consider policy learning methods given by the following scores for varying $c$,
$$\Gamma_i=\frac{1-\hat\phi^2(X_i)}{c+1-\hat\phi^2(X_i)}\prns{\hat\mu(+\mid X_i)-\hat\mu(-\mid X_i)+(Y_i-\hat\mu(A_i\mid X_i))/(A_i+\hat\phi(X_i))}.$$
\Cref{fig:synthetic_pad} shows the results when running 50000 replicates for each of 33 $c$ values, evenly spaced on the logarithmic scale of the plot axis, with $n=2000$ and $\beta=3.5$. In the well-specified scenarios, full retargeting ($c\to\infty$) is best while in the misspecified scenarios, full retargeting is not bad but some regularization may offer minimal improvements.

These minimal improvements are given by the best regularization parameter. In practice, $c$ must be chosen, such as using cross validation (where, \eg, out-of-sample estimates of policy value are estimated using some efficient estimator). Therefore these improvements may be too minimal to consider, given that choosing the regularization parameter $c$ is non-trivial.
Indeed, as discussed in \cref{sec:discussion misspecification and shift}, optimal retargeting may be both safest and most effective even under misspecification. Nonetheless, bias regularization may provide another knob to control retargeting that may be useful in some applications.
}

\section{Omitted Proofs}\label{sec:omittedproofs}

\begin{proof}[Proof of \cref{lemma:weightinvariance}]
Since $\Pi$ allows for unconstrained choice of $\pi(\cdot\mid X)$ across values of $X$, we must have
\begin{align*}\argmax_{\pi\in\Pi} R(\pi;w,\rho)
&=\braces{\pi\in\Pi:
\pi(\cdot\mid X)\in\argmax_{p\in\DeltaA}w(X)\sumA(p(a)-\rho(a\mid X))\mu(a\mid X),\ \text{a.s.}
}\\
&=\braces{\pi\in\Pi:
\pi(\cdot\mid X)\in\argmax_{p\in\DeltaA}\sumA p(a)\mu(a\mid X),\ \text{a.s.}
}
\\&=\Pi^*.\end{align*}

Let $\pi\in\Pi^*\cap\Pi_0$ and $\pi'\in\Pi_0$. Then, since $\pi\in\Pi^*$, by the above we must have $R(\pi;w,\rho)\leq R(\pi';w,\rho)$. Since $\pi'\in\Pi_0$ was arbitrary, we have $\pi\in\Pi_0^*$. 

Let $\pi\in\Pi_0^*$ and $\pi'\in\Pi$. By assumption, there exists $\pi''\in\Pi^*\cap\Pi_0\neq\varnothing$. Because $\pi''\in\Pi_0$, we have 
$R(\pi;w,\rho)\leq R(\pi'';w,\rho)$. Because $\pi''\in\Pi^*$ and by the above, we have $R(\pi'';w,\rho)\leq R(\pi';w,\rho)$. Hence, $R(\pi;w,\rho)\leq R(\pi';w,\rho)$. Since $\pi'\in\Pi$ was arbitrary, we have $\pi\in\Pi^*$. Moreover, by definition, $\pi\in\Pi_0$. Hence, $\pi\in\Pi^*\cap\Pi_0$. 
\end{proof}

{\blockedit
\begin{proof}[Proof of \cref{lemma:weightinvariance2}]
We first note that $\Pi^*\subseteq \Pi^*_n$ almost surely because any $\pi^*\in\Pi^*$ must be point-wise optimal at almost every $x$.
Therefore, $\Pi^*\cap\Pi_0\neq\varnothing$ implies that $\Pi_n^*\cap\Pi_0\neq\varnothing$. The proof then proceeds exactly as that of \cref{lemma:weightinvariance}.
\end{proof}
}

{\blockedit
\begin{proof}[Proof of \cref{lemma:unifconv}]
The quantity to be bounded is:
$$
\mathfrak D_n=\sup_{\pi,\pi'\in\Pi}\abs{(\tilde V_n(\pi;w)-\tilde V_n(\pi';w))-(V_n(\pi;w)-V_n(\pi';w))}.
$$

Define $\pi(X_i)=(\pi(a\mid X_i))_{a\in\A}\in\R{\A}$ and $\Gamma_i\in\R{\A}$ such that 
$$\Gamma_{i,a}=\indic{A_i=a}w(X_i){\phi^{-1}(A_i\mid X_i)}(Y_i-\mu(A_i\mid X_i)).$$
With this notation, notice that we have
$\tilde V_n(\pi;w)-V_n(\pi;w)=\frac1n\sum_{i=1}^n\ip{\Gamma_i}{\pi(X_i)}$. Moreover, notice that $\Eb{\Gamma_{i,a}\mid X_i}=0$ whence $\E\ip{\Gamma_i}{\pi(X_i)}=0$.

By boundedness assumptions, there is some $B>0$ such that $\abs{\ip{\Gamma_i}{\pi(X_i)-\pi'(X_i)}}\leq \abs{{\phi^{-1}(A_i\mid X_i)}(Y_i-\mu(A_i\mid X_i))}<B$. Then, the final statement of Theorem 2.3 of \citet{bousquet2002bennett} (simplified using $\sqrt{x+y}\leq\sqrt{x}+\sqrt{y}$ for $x,y\geq0$) yields that with probability at least $1-\delta$,
\begin{align}\label{eq:bennetineq}
\mathfrak D_n&\leq
\E\mathfrak D_n+
\sqrt{\frac{2\log(1/\delta)}n}\sup_{\pi,\pi'\in\Pi_0}\Eb{\ip{\Gamma_i}{\pi(X_i)-\pi'(X_i)}^2}^{1/2}
\\\notag&\phantom{\leq}+\frac{2B\sqrt{\log(1/\delta)}}{n}\sqrt{\E\mathfrak D_n}+\frac{B\log(1/\delta)}{3n}.
\end{align}
Note that
\begin{align}\label{eq:boundsupvaraince}
\sup_{\pi,\pi'\in\Pi_0}&\Eb{\ip{\Gamma_i}{\pi(X_i)-\pi'(X_i)}^2}^{1/2}
\leq 2\sup_{\pi\in\Pi}\Eb{\ip{\Gamma_i}{\pi(X_i)-\rho(X_i)}^2}^{1/2}
\\\notag&=2\Eb{\prns{w(X_i)\frac{\pi(A_i\mid X_i)-\rho(A_i\mid X_i)}{\phi(A_i\mid X_i)}(Y_i-\mu(A_i\mid X_i))}^2}^{1/2}
\\\notag&=2\Eb{w^2(X_i)\sumA\frac{\prns{\pi(a\mid X_i)-\rho(a\mid X_i)}^2}{\phi(a\mid X_i)}\sigma^2(a\mid X_i)}^{1/2}
\\\notag&=2\sqrt{\Omega(w,\rho)}.
\end{align}
We next proceed to bound $\E\mathfrak D_n$.

Define the Rademacher complexity
$$
\mathfrak R_n=\frac{1}{2^n}\sum_{\epsilon\in\{-1,+1\}^n}
\E\sup_{\pi,\pi'\in\Pi_0}\abs{\frac1n
\sum_{i=1}^n\epsilon_i\ip{\Gamma_i}{\pi(X_i)-\pi'(X_i)}
}.
$$
Letting $(\Gamma'_i,X'_i)\sim(\Gamma_i,X_i)$ be an iid copy (ghost sample), a standard symmetrization argument yields
\begin{align*}
\E\mathfrak D_n&=\E\sup_{\pi,\pi'\in\Pi}\abs{\frac1n\sum_{i=1}^n\ip{\Gamma_i}{\pi(X_i)-\pi'(X_i)}}
\\
&=\E\sup_{\pi,\pi'\in\Pi}\abs{\Eb{\frac1n\sum_{i=1}^n\prns{\ip{\Gamma_i}{\pi(X_i)-\pi'(X_i)}-\ip{\Gamma'_i}{\pi(X'_i)-\pi'(X'_i)}}
\mid \Gamma_{1:n},X_{1:n}
}}
\\
&\leq\E\sup_{\pi,\pi'\in\Pi}\abs{\frac1n\sum_{i=1}^n\prns{\ip{\Gamma_i}{\pi(X_i)-\pi'(X_i)}-\ip{\Gamma'_i}{\pi(X'_i)-\pi'(X'_i)}
}}
\\
&=\frac{1}{2^n}\sum_{\epsilon\in\{-1,+1\}^n}\E\sup_{\pi,\pi'\in\Pi}\abs{\frac1n\sum_{i=1}^n\epsilon_i\prns{\ip{\Gamma_i}{\pi(X_i)-\pi'(X_i)}-\ip{\Gamma'_i}{\pi(X'_i)-\pi'(X'_i)}
}}
\\
&\leq\frac{2}{2^n}\sum_{\epsilon\in\{-1,+1\}^n}\E\sup_{\pi,\pi'\in\Pi}\abs{\frac1n\sum_{i=1}^n\epsilon_i{\ip{\Gamma_i}{\pi(X_i)-\pi'(X_i)}
}}
\\&=2\mathfrak R_n.
\end{align*}

Finally, 
Theorem 1 of \citet{zhou2018offline} yields
$$
\mathfrak R_n
\leq 27.2\sqrt{2}\prns{\kappa(\Pi_0)+8}n^{-1/2}
\sup_{\pi,\pi'\in\Pi_0}\Eb{\ip{\Gamma_i}{\pi(X_i)-\pi'(X_i)}^2}^{1/2}+o(n^{-1/2}).
$$
Combining with \cref{eq:bennetineq,eq:boundsupvaraince} and $\E\mathfrak D_n\leq2\mathfrak R_n$ completes the proof.
\end{proof}
}

\begin{proof}[Proof of \cref{lemma:omegaform}]
Since $\pi\in\Pi$ is unconstrained we have
$$
\Omega(w,\rho)=
\Eb{
\sup_{p\in\Delta^{\A}}
w^2(X)\sumA\frac{\sigma^2(a\mid X)}{\phi(a\mid X)}\prns{p(a)-\rho(a\mid X)}^2
}.
$$
Consider the inner optimization for every $X$. Note that it is the sup of a convex function in $p$ over a convex region. Therefore, an optimal solution occurs at an extreme point. Hence, using $1^2=1$,
\begin{align*}
\Omega(w,\rho)&=
\Eb{
\max_{p\in\{0,1\}^{\A}\cap\Delta^{\A}}
w^2(X)\sumA\frac{\sigma^2(a\mid X)}{\phi(a\mid X)}\prns{p(a)-\rho(a\mid X)}^2
}\\
&=
\Eb{
\max_{p\in\Delta^{\A}}
w^2(X)\sumA\frac{\sigma^2(a\mid X)}{\phi(a\mid X)}
\prns{p(a)-2p(a)\rho(a\mid X)+\rho^2(a\mid X)}
}\\
&=
\Eb{
w^2(X)\prns{
\sumA\frac{\sigma^2(a\mid X)}{\phi(a\mid X)}\rho^2(a\mid X)
+
\max_{p\in\Delta^{\A}}
\sumA\frac{\sigma^2(a\mid X)}{\phi(a\mid X)}
\prns{1-2\rho(a\mid X)}p(a)
}}.
\end{align*}
Taking the maximum over $p$ picks out the largest coefficient, giving the result.
\end{proof}

\begin{proof}[Proof of \cref{lemma:twoarmomega}]
From \cref{lemma:omegaform}, since $\phi(\pm\mid x)=(1\pm\phi(x))/2$, we have
\begin{align*}
\Omega(w,\rho)=\E\Biggl[&
2w^2(X)\Biggl(
\frac{\sigma^2(+\mid X)}{1+\phi(X)}\rho^2(+\mid X)
+
\frac{\sigma^2(-\mid X)}{1-\phi(X)}\rho^2(-\mid X)
\\&\phantom{=}+
\max\braces{
\frac{\sigma^2(+\mid X)}{1+\phi(X)}(1-2\rho(+\mid X)),\ 
\frac{\sigma^2(-\mid X)}{1-\phi(X)}(1-2\rho(-\mid X))
}
\Biggr)
\Biggr].
\end{align*}
Fix $w$ and consider $\min_\rho\Omega(w,\rho)$. Since $\rho$ is unconstrained, $\rho(\cdot\mid X)$ minimizes the inside of the above expectation for each $X$. Fix $x$.
Let $\beta_{\pm}=\frac{\sigma^2(\pm\mid x)}{1\pm\phi(x)}\geq0$.
Then $\rho(\cdot\mid x)$ solves
$\min_{\rho(\pm\mid x)}f(\rho(+\mid x),\rho(-\mid x))$, where
$$f(z_+,z_-)=\beta_{+}z_+^2+\beta_{-}z_-^2+\max\braces{
\beta_{+}(1-2z_+),\ \beta_{-}(1-2z_-)
}.$$
We will show that $z_\pm=\frac12$ is an optimal solution.
If either $\beta_{+}=0$ or $\beta_{-}=0$ then $z_-=z_+=\frac12$ is clearly an optimizer. Suppose $\beta_\pm>0$.
$f(z_+,z_-)$ is convex and is therefore optimized at critical points.
To enumerate the critical points, first consider the case where
$\beta_{+}(1-2z_+)>\beta_{-}(1-2z_-)$.
Then $f(z_+,z_-)$ is differentiable and a critical point must satisfy
$\partial_{z_+}f(z_+,z_-)=2(\beta_+z_+-\beta_+)=0\implies z_+=1$,
$\partial_{z_-}f(z_+,z_-)=2\beta_-z_-=0\implies z_-=0$,
which contradicts $\beta_{+}(1-2z_+)>\beta_{-}(1-2z_-)$, so there are no such critical points. The same holds for the case $\beta_{+}(1-2z_+)<\beta_{-}(1-2z_-)$. Consider the case $\beta_{+}(1-2z_+)=\beta_{-}(1-2z_-)$. Then $z_+=\frac12+\frac{\beta_-}{\beta_+}\prns{z_--\frac12}$ and $f(z_+,z_-)=\frac{\beta_-+\beta_+}{4\beta_+}(\beta_++\beta_-(2z_--1)^2)$, which is optimized at $z_-=\frac12$, leading to $z_+=\frac12$. So $\rho_0(\cdot\mid x)=\frac12$ is optimal.

For $\rho=\rho_0$, we have
\begin{align*}
\Omega(w,\rho_0)=\frac12\Eb{
w^2(X)\prns{
\frac{\sigma^2(+\mid X)}{1+\phi(X)}
+
\frac{\sigma^2(-\mid X)}{1-\phi(X)}
}
}.
\end{align*}
We next optimize this over $w$ subject to $\E w(X)=1$. 
Let $\kappa(x)=\prns{
\frac{\sigma^2(+\mid X)}{1+\phi(X)}
+
\frac{\sigma^2(-\mid X)}{1-\phi(X)}
}$.
Since the above is convex in $w$, we must have that the optimal $w$ satisfies
$$
\exists \lambda\in\R\;:\;\Eb{w(X)}=1,\,
w(x)\kappa(x)=\lambda.
$$
Therefore the optimal $w$ satisfies
$w(x)\propto\kappa^{-1}(x)$. 
\end{proof}

\begin{proof}[Proof of \cref{lemma:multiarmomega}]
From \cref{lemma:omegaform},
\begin{align*}
\Omega(w,\rho)&=
\Eb{
w^2(X)
\prns{
\sumA{\zeta(a\mid X)}\rho^2(a\mid X)
+\max_{a\in\A}{\zeta(a\mid X)}(1-2\rho(a\mid X))
}
}.
\end{align*}
Therefore, since $\rho$ is unconstrained $\rho(\cdot\mid x)$ solves $\min_{\rho(\cdot\mid x)}f((\rho(a\mid x))_{a\in\A})$ where
$$f(z)=\sumA\zeta(a\mid x)z_a^2+\max_{a\in\A}\zeta(a\mid x)(1-2z_a).$$
Fix $x$. Then we can write this \edit{convex} optimization problem as:
\begin{align*}
\min_zf(z)\ =\ \min_{z,y}&\ \sumA\zeta(a\mid x)z_a^2+y\\
\text{s.t.}&\ y\geq \zeta(a\mid x)(1-2z_a)\;\forall a\in\A
\end{align*}
The Lagrangian of this problem, for a dual variable $\lambda\in\R{\A}_+$, is 
\begin{align*}L(z,y,\lambda)&=\sumA\zeta(a\mid x)z_a^2+y+\sumA\lambda_a(\zeta(a\mid x)(1-2z_a)-y)\\&=\sumA \zeta(a\mid x)(z_a^2-2\lambda_az_a)+y(1-\sumA\lambda_a)+\sumA\zeta(a\mid x)\lambda_a.\end{align*}
The KKT conditions for $(z,y,\lambda)$ to be a primal-dual optimal are that
\begin{align*}
y&\geq \zeta(a\mid x)(1-2z_a)\;\forall a\in\A&&\text{(Primal feasible)}\\
\lambda&\in\R{\A}_+&&\text{(Dual feasible)}\\
0&=2\zeta(a\mid x)(z_a-\lambda_a)&&\text{($z$ stationarity)}\\
0&=1-\sumA\lambda_a&&\text{($y$ stationarity)}\\
0&=\lambda_a(\zeta(a\mid x)(1-2z_a)-y)&&\text{(Complementary slackness)}
\end{align*}
{\blockedit
Now consider the solution given by
$$
y=\frac{m-2}{\sumA \zeta^{-1}(a\mid x)},\quad
\lambda_a=z_a=\frac12\prns{1-\frac{y}{\zeta(a\mid x)}}.
$$
It can be directly validated that this solution satisfies all of the above KKT conditions via plugging in the solution and some algebra. Since the problem is convex we conclude this is an optimal solution.

Next, note that for the above solution we have that the objective, which is optimal, is
$$
\sumA\zeta(a\mid x)z_a^2+y=\frac14\kappa(x),
$$
and therefore
$$\Omega(w,\rho_0)=\frac14\Eb{w^2(X)\kappa(X)}.$$
Repeating the argument from the proof of \cref{lemma:twoarmomega}, we have that $w_0\propto \kappa^{-1}(x)$ is optimal.}
\end{proof}

\begin{proof}[Proof of \cref{lemma:biasworstcase}]
Since the sign of $\mu(a\mid x)$ is completely unconstrained across $a,x$, we have
\begin{align*}
B(w;\lambda)&=\sup_{\pi\in\Pi,\ \fmagd{\mu_{\max}}_{L_2}\leq\lambda}\abs{\Eb{(w(X)-1)\sum_{a\in\A}\pi(a\mid X)\mu(a\mid X)}}\\
&=\sup_{\pi\in\Pi,\fmagd{\mu_{\max}}_{L_2}\leq1}\Eb{(w(X)-1)\sumA\pi(a\mid X)\abs{\mu(a\mid X)}}\\
&=\sup_{\fmagd{\mu_{\max}}_{L_2}\leq1}\Eb{(w(X)-1)\mu_{\max}(X)},
\end{align*}
which yields the result by the self-duality of $L_2$.
\end{proof}

\begin{proof}[Proof of \cref{lemma:biasregularization}]
From \cref{lemma:multiarmomega}, we have that $\inf_\rho\Omega(w,\rho)=\frac14\Eb{w^2(X)\kappa(X)}$. Therefore, $$\mathcal E^2_\lambda(w)=
\Eb{\frac{1}{4}\kappa(X)w^2(X)+\lambda^2(w(X)-1)^2}.$$
Since this is convex in $w$, we must have that the optimal $w$ satisfies
$$
\exists a\in\R\;:\;\Eb{w(X)}=1,\,\frac12\kappa(x)w(x)+2\lambda^2(w(x)-1)=a.
$$
Therefore, the optimal $w$ satisfies $w(x)\propto(\kappa(x)+4\lambda^2)^{-1}$.
\end{proof}

{\blockedit
\begin{proof}[Proof of \cref{lemma:unifconv2}]
The proof is the same as that of \cref{lemma:unifconv} with $\Gamma_{i,a}=\mu(a\mid X_i)$.
\end{proof}

\begin{proof}[Proof of \cref{lemma:finitepolicyconvergence}]
Since $\Pi_0\cap\Pi^*\neq\varnothing$, we have $\max_{\pi\in\Pi_0}V_n(\pi;w)=\frac1n\sum_{i=1}^nw(X_i)\max_{a\in\A}\mu(a\mid X_i)$.
Therefore, given $w>0$,
\begin{align*}
&\pi\in\argmax_{\pi\in\Pi_0}V_n(\pi;w)\\
&\iff
\frac1n\sum_{i=1}^nw(X_i)\sumA\pi(a\mid X_i)\prns{\max_{a'\in\A}\mu(a'\mid X_i)-\mu(a\mid X_i)}=0\\
&\iff
\pi(a\mid X_i)\prns{\max_{a'\in\A}\mu(a'\mid X_i)-\mu(a\mid X_i)}=0~\forall a\in\A,\,i=1,\dots,n,
\end{align*}
the probability of which is precisely $p_\pi^n$.
Recall that $p_\pi=1$ if and only if $\pi\in\Pi^*$, whence by union bound over $\Pi^*\cap\Pi_0=\{\pi\in\Pi_0:p_\pi\neq1\}$ we have the result.
\end{proof}

\begin{proof}[Proof of \cref{lemma:finitepolicyconvergence2}]
First note that if $\sigma^2(A\mid X)=0$ almost surely the result is trivial since then $\tilde R_n(\pi;w)=R_n(\pi;w)$ for all $w$. So suppose $\sigma^2(A\mid X)>0$ with positive probability. Then $\Omega_{\min}=\inf_w\Omega(w,\rho)>0$. 

Recall that $\tilde R_n(\pi;w,\rho)$ has the exact same minimizers as $\tilde V_n(\pi;w)$ and similarly that $R_n(\pi;w,\rho)$ has the exact same minimizers as $V_n(\pi;w)$. We therefore proceed to consider the statement about the set of maximizers of the former of each.

Recall that $\tilde R_n(\pi;w,\rho)-R_n(\pi;w,\rho)=\frac1n\sum_{i=1}^nw(X_i)\frac{\pi(A_i\mid X_i)-\rho(A_i\mid X_i)}{\phi(A_i\mid X_i)}\prns{Y_i-\mu(A_i\mid X_i)}$.
Let $s^2_i(\pi)=w^2(X_i)\sumA\frac{\sigma^2(a\mid X_i)}{\phi(a\mid X_i)}\prns{\pi(a\mid X_i)-\rho(a\mid X_i)}^2$. Then by boundedness, for some $B>0$, Bernstein's inequality yields
\begin{align*}
\Prb{\abs{\tilde R_n(\pi;w,\rho)-R_n(\pi;w,\rho)}>t\mid X_{1:n}}
&\leq 2\exp\prns{\frac{-nt^2}{\frac2n\sum_{i=1}^ns^2_i(\pi)+Bt}}.
\end{align*}
Note that $\gamma_n(w)=\min_{\pi\in\Pi_0} R_n(\pi;w,\rho)-\min_{\pi\in\Pi_0\backslash\Pi_{0,n}^*(w)} R_n(\pi;w,\rho)$. Hence, by union bound,
$$\Prb{\argmax_{\pi\in\Pi_0}\tilde V_n(\pi;w)\not\subseteq\Pi_{0,n^*}(w)\mid X_{1:n}}
\leq
2\abs{\Pi_0}\exp\prns{\frac{-n\gamma_n^2(w)}{\frac8n\sum_{i=1}^ns^2_i(\pi)+2B\gamma_n(w)}}.
$$

Let $s^2(\pi)=\E s^2_i(\pi)$. Boundedness and
Hoeffding's inequality gives that\break $\Prb{\frac1n\sum_{i=1}^ns^2_i(\pi)>s^2(\pi)+\Omega_{\min}}\leq\exp(-\eta n\Omega_{\min}^2)$ for some $\eta$. By union bound, with probability at least $1-\abs{\Pi_0}\exp(-n\eta\Omega_{\min}^2)$, we have $\sup_{\pi\in\Pi_0}\frac1n\sum_{i=1}^ns^2_i(\pi)\leq \sup_{\pi\in\Pi_0}s^2(\pi)+\Omega_{\min}\leq 2\Omega(w,\rho)$. Choosing $\alpha=(8\Omega_{\min}/B)^2$ yields the result.
\end{proof}
}

\end{document}